\documentclass[acmsmall]{acmart}
\AtBeginDocument{%
  }

\acmJournal{TACO}

\usepackage{algpseudocode}
\usepackage[linesnumbered,ruled,vlined]{algorithm2e}
\usepackage{pifont}
\usepackage{graphicx}
\usepackage{textcomp}
\usepackage{xcolor}
\usepackage{pgf}
\usepackage{tikz}
\usepackage{color}
\usepackage{stfloats}
\usepackage{multirow}
\usepackage{diagbox}

\usepackage{booktabs}
\usepackage{arydshln}

\usepackage{lipsum}
\usepackage{wrapfig}
\usepackage[normalem]{ulem}

\newcommand{\squishlist}{
   \begin{list}{$\bullet$}
    { \setlength{\itemsep}{0pt}      \setlength{\parsep}{0pt}
      \setlength{\topsep}{3pt}       \setlength{\partopsep}{0pt}
      \setlength{\listparindent}{-2pt}
      \setlength{\itemindent}{-5pt}
      \setlength{\leftmargin}{1em} \setlength{\labelwidth}{0em}
      \setlength{\labelsep}{0.5em} } }

\newcommand{\squishend}{
    \end{list}  }

\newcommand*\blackcircledempty[1]{\tikz[baseline=(char.base)]{
        \node[shape=circle, text={rgb,255:red,0;green,0;blue,0}, font=\small, draw={rgb,255:red,0;green,0;blue,0},inner sep=0.5pt] (char) {#1};}}

\newcommand{\marknumber}[1]{%
  \tikz[baseline=(char.base)]{
    \node[shape=circle,draw,minimum size=0.5em,inner sep=0pt,fill=black,text=white,text width=0.7em,align=center,baseline=-0.5em] (char) {\scriptsize #1};
  }%
}

\newcommand{\closerxrightarrow}[2][]{\mathrel{\mathop{\xrightarrow{#1}}\limits^{\raisebox{-1.5pt}[0pt][0pt]{\scriptsize #2}}}}

\begin{document}

\title{Characterizing and Understanding HGNN Training on GPUs}

\author{Dengke Han}
\email{handengke21s@ict.ac.cn}
\orcid{0000-0003-0641-5779}
\author{Mingyu Yan}
\authornote{Corresponding author.}
\email{yanmingyu@ict.ac.cn}
\orcid{0000-0002-6915-955X}
\author{Xiaochun Ye}
\email{yexiaochun@ict.ac.cn}
\orcid{0000-0003-4598-1685}
\author{Dongrui Fan}
\email{fandr@ict.ac.cn}
\orcid{0000-0001-5219-0908}
\orcid{}
\affiliation{%
  \institution{State Key Laboratory of Processors, Institute of Computing Technology,
Chinese Academy of Sciences and University of Chinese Academy of Sciences}
  \streetaddress{6 Kexueyuan Nanlu, Zhongguancun, Haidian, Beijing, China}
  \city{Beijing}
  \state{Beijing}
  \country{China}
  \postcode{100190}
}

\renewcommand{\shortauthors}{D. Han et al.}

\begin{abstract}

Owing to their remarkable representation capabilities for heterogeneous graph data, Heterogeneous Graph Neural Networks (HGNNs) have been widely adopted in many critical real-world domains such as recommendation systems and medical analysis. Prior to their practical application, identifying the optimal HGNN model parameters tailored to specific tasks through extensive training is a time-consuming and costly process. 
To enhance the efficiency of HGNN training, it is essential to characterize and analyze the execution semantics and patterns within the training process to identify performance bottlenecks. In this study, we conduct a comprehensive quantification and in-depth analysis of two mainstream HGNN training scenarios, including single-GPU and multi-GPU distributed training. Based on the characterization results, we reveal the performance bottlenecks and their underlying causes in different HGNN training scenarios and propose optimization guidelines from both software and hardware perspectives.

\end{abstract}

\ccsdesc[500]{General and reference~General literature}
\ccsdesc[500]{Computer systems organization~Architectures}
\ccsdesc[500]{Theory of computation~Graph algorithms analysis}
\ccsdesc[300]{Computing methodologies~Artificial intelligence}

\keywords{Heterogeneous Graph Neural Networks, Graph Neural Networks Training, Characterization, Quantitative Analysis, Optimization Guidelines}


\maketitle

\section{Introduction}
\label{sec:introduction}

In recent years, Graph Neural Networks (GNNs) have demonstrated exceptional capabilities in representing and processing graph data in non-Euclidean spaces~\cite{comprehensive_gnn_survey, gnn1, gnn2, gnn3}. The early successes of GNNs are predominantly in the domain of homogeneous graphs (HomoGs), characterized by a single type of entity and adjacency relation~\cite{GCN,GAT,GraphSage}. However, many real-world data in complex systems are more aptly represented as heterogeneous graphs (HetGs) which consist of multiple types of entities and relations embodied by various types of vertices and edges, respectively. In contrast to HomoGs, HetGs contain not only the structural information inherent in graph data but also the rich semantic information embedded in the relations~\cite{HG_survey, hgnn_survey_shichuan}. Due to the powerful representation ability of HetGs, Heterogeneous Graph Neural Networks (HGNNs) have been developed and widely adopted in many critical fields including recommendation systems~\cite{weibo-recommendation, dataset-recommendation, hierarchy-recommendation}, cybersecurity~\cite{malware-detection1, tencent_malware_detection, abnormal_event_detection}, medical analysis~\cite{medical_analysis, medical_analysis_aaai}, traffic prediction~\cite{traffic-demand-forecasting,traffic-speed-prediction,traffic-forecasting} and many others.


Unlike GNNs, which recursively aggregate the feature vectors of neighboring vertices~\cite{GCN,comprehensive_gnn_survey,Comprehensive_Survey_GNN_Distributed_Training} to obtain structural information in HomoGs, HGNNs employ a different execution semantics to extract both structural and semantic information. Specifically, most of the mainstream HGNN models can be partitioned into four primary execution stages~\cite{R-GCN,R-GAT,HAN,Simple-HGN,SeHGNN}: \blackcircledempty{1} \textit{Semantic Graph Build} (SGB); \blackcircledempty{2} \textit{Feature Projection} (FP); \blackcircledempty{3} \textit{Neighbor Aggregation} (NA); \blackcircledempty{4} \textit{Semantic Fusion} (SF). The SGB stage partitions the original HetGs into multiple semantic graphs. The FP and NA stages perform conventional GNN processes, operating independently within each semantic graph. Subsequently, the SF stage fuses the results of the NA stage across different semantic graphs.


Applying HGNNs to downstream tasks such as vertex classification or link prediction requires a comprehensive training process to find the optimal model parameters. Inherited from conventional GNNs, training HGNNs on computing nodes like GPUs primarily involves two methods: full-batch and mini-batch training. Unlike full-batch training on the entire graph, mini-batch training iteratively trains on mini-batches generated from a group of target vertices, significantly reducing memory footprints and speeding up convergence~\cite{GraphSage}. Moreover, distributed training across multiple computing nodes distributes the training load, improving the overall performance of the training process. Although the introduction of mini-batch training and distributed training solutions has improved the scalability and training efficiency of GNN models respectively, the training process itself remains extremely time-consuming and resource-intensive~\cite{training_expensive1, training_expensive2}. HGNNs, due to their higher algorithmic complexity compared to GNNs~\cite{hgnn_survey_hgnn}, exhibit even higher costs than GNNs.

Quantitatively analyzing and characterizing the execution behaviors and patterns of HGNN models is crucial for enhancing their execution efficiency. Currently, extensive efforts have been dedicated to characterizing GNNs~\cite{understand_GCN, understand_distributed_gnn, training_expensive2, gnn_analysis_of_bottlenecks, understand_GNN}, contributing to a substantial understanding within the research community of their execution patterns. However, there is limited effort focused on HGNN characterization. On one hand, GNNs and HGNNs exhibit distinct execution semantics and patterns, with HGNNs demonstrating more intricate execution behaviors. Consequently, characterization results for GNNs cannot be directly applied to optimize HGNN efficiency. On the other hand, existing characterization study on HGNNs~\cite{understand_HGNN, MetaNMP, HiHGNN} have primarily focused on inference tasks. Yet, the training process is significantly more complex than inference, rendering existing inference-focused characterizations insufficient for guiding training efficiency optimizations. 

To address these gaps, we conduct a comprehensive characterization and quantitative analysis of HGNN training across single-node and multi-node distributed platforms, incorporating both full-batch and mini-batch training methods. Specifically, we undertake a quantitative characterization from various perspectives, encompassing the intrinsic characteristics of the HGNN execution stages, a comparative analysis of forward and backward propagation, the overall attributes of the training process, and the influence of metapath properties on training performance. Furthermore, a comparative analysis of HGNN and GNN training is carried out. Finally, we propose optimization guidelines from both software and hardware perspectives. We systematically summarize our 22 findings based on the four aforementioned characterization perspectives as follows:

\squishlist

\item \textit{Execution Characteristics of HGNN Models}: (1) The NA stage dominates the most execution time both in forward and backward propagation, which is considered the predominant stage across all the execution stages of HGNNs; (2) There exists a distinct hybrid execution pattern between HGNN stages, each characterized by unique execution bounds, resulting in varying demands on hardware resources.; (3) Mini-batch sampling accounts for the majority of the execution time in each epoch of mini-batch training in both single-GPU and multi-GPU distributed scenarios.

\item \textit{Comparison between Backward and Forward Propagation}: (1) Forward propagation is generally more time-consuming than backward propagation; (2) Although the kernels used during backward propagation are similar to those in forward propagation, there are significant differences in time distribution across kernel types and in the execution characteristics of the same kernels; (3) Backward propagation requires greater memory access and demonstrates lower data locality compared to the forward pass. However, stages of execution that involve a high number of additions are expected to have a lower computational workload during backward propagation.

\item \textit{Overall Analysis of End-to-end Training Process}: (1) In comparison to pure inference, the training process necessitates greater memory footprint, with the majority memory allocation lies on the NA stage; (2) The vast majority of kernel stalls during the training process originate from memory dependency except for \textit{sgemm} kernel; (3) The primary factor influencing the performance improvement of multi-GPU distributed training is the contention for shared hardware resources.

\item \textit{Exploration Related to Metapaths}: (1) An increase in both the length and number of metapaths significantly extends execution time during the NA stage, while the FP and SF stages are sensitive only to variations in the number of metapaths. (2) For large-scale datasets, an increase in both the length and number of metapaths leads to a substantial rise in sampling time. While sampling time is primarily influenced by changes in the number of metapaths for small datasets.

\squishend

\section{Background}
\label{sec:background}


\subsection{Heterogeneous Graphs and Semantic Graphs}

\begin{figure*}[!ht] 
	\centering
	\includegraphics[width=0.96\textwidth]{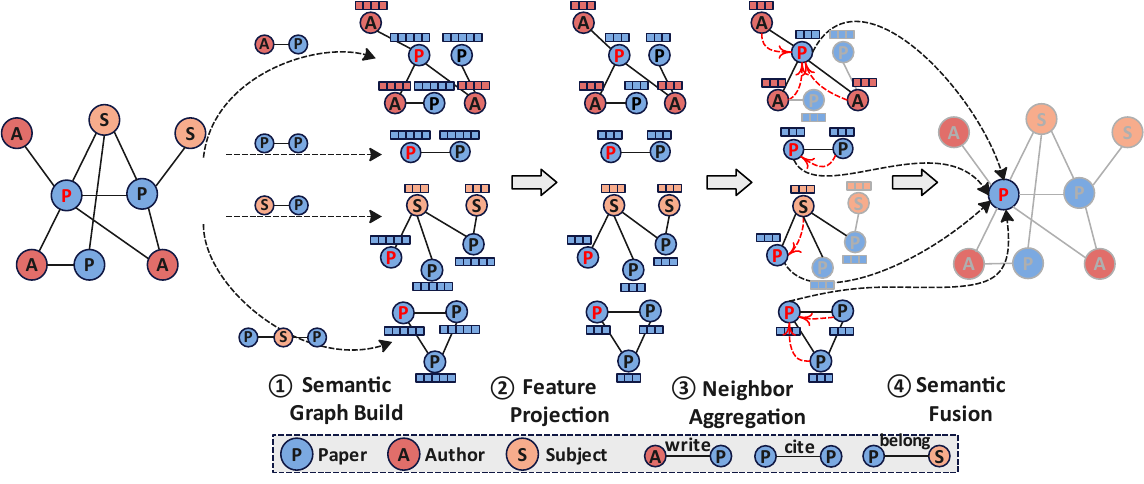}
	\caption{Illustration of HetGs and HGNNs.}
	\label{fig:background_hgnns}
\end{figure*}

Fig.~\ref{fig:background_hgnns} illustrates a simple example of a HetG from the ACM dataset, which includes three types of vertices, A (Author), P (Paper), and S (Subject), along with three types of adjacency relations between them: author$\closerxrightarrow{\rm writes}$paper, paper$\closerxrightarrow{\rm cites}$paper, and paper$\closerxrightarrow{\rm belongs\ to}$subject (abbreviated as AP, PP and PS). Moreover, the inverse of these relations also holds significant meaning like PA and SP. In addition to direct relations like AP, different combinations of these direct relations can form higher-order relations, referred to as metapaths. For instance, in Fig.~\ref{fig:background_hgnns}, PSP represents a metapath composed of PS and SP, signifying that two papers are linked by a shared subject, implying a strong likelihood that the two papers pertain to the same research area. Each type of relation or metapath represents unique semantic information between the two endpoints connected.

Semantic graphs are derived from the original HetG by focusing on specific relations or metapaths. Each semantic graph is constructed to represent only a single type of relation or metapath. As illustrated in Fig.~\ref{fig:background_hgnns}, semantic graphs are constructed based on the AP, PP, PS, and PSP, respectively. This process of segmentation allows for the separation of multiple semantic aspects within the original HetG, thereby facilitating more efficient information extraction. Numerous HGNN models~\cite{HAN,MAGNN,SeHGNN} utilize metapaths to construct semantic graphs, while others~\cite{R-GCN,R-GAT} employ direct relations, i.e., edge types, for semantic graph construction.

\subsection{Heterogeneous Graph Neural Networks}

Given the inefficiencies of traditional GNNs in extracting semantic information from HetGs, a multitude of specialized HGNNs have recently emerged~\cite{HAN, R-GCN, R-GAT, MAGNN, Simple-HGN}, which incorporate both structural and semantic information from HetGs through distinct neighborhood aggregation and semantic fusion schemes. Specifically, these schemes involve separate aggregation of neighbors with different semantics followed by their fusion. In general, most prevalent HGNNs usually contain four major execution stages as shown in Fig.~\ref{fig:background_hgnns}: \blackcircledempty{1} \textit{Semantic Graph Build} stage partitions the original HetG into several semantic graphs; \blackcircledempty{2} \textit{Feature Projection} stage transforms the feature vectors of vertices in each semantic graph to new ones using multi-layer perceptrons; \blackcircledempty{3} \textit{Neighbor Aggregation} stage aggregates features from neighbors for each target vertex within semantic graphs; \blackcircledempty{4} \textit{Semantic Fusion} stage fuses the results of the NA stage across different semantic graphs for each vertex to aggregate the semantic information.

\subsection{HGNN Training}

\begin{figure*}[!ht]
	\centering
	\includegraphics[width=0.96\textwidth]{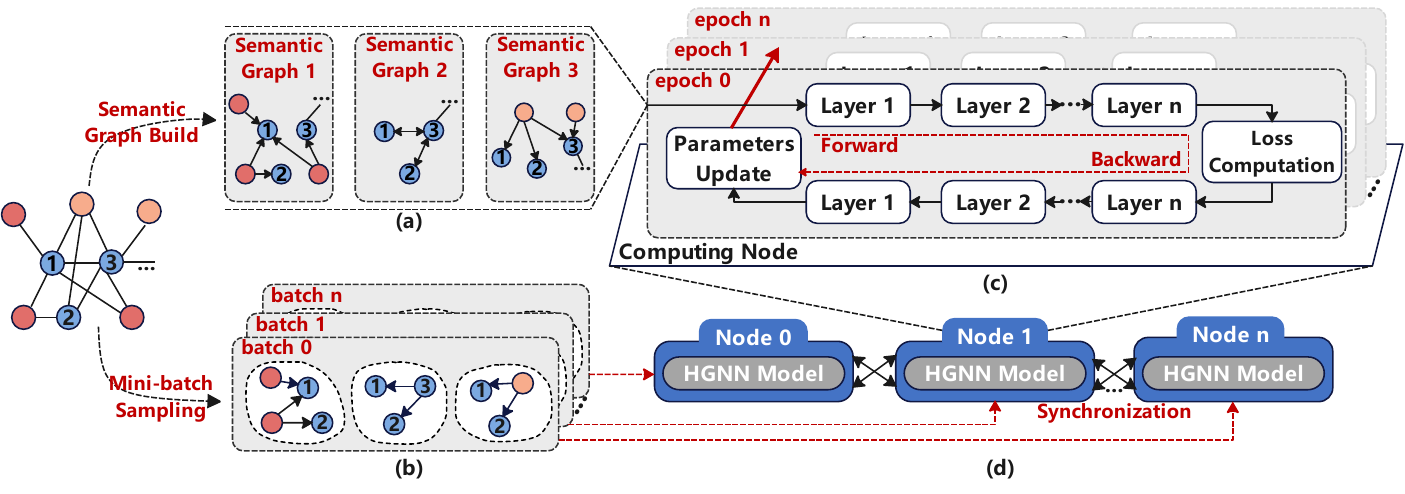}
	\caption{Illustration of HGNN training: (a) SGB stage; (b) Mini-batch sampling process; (c) Training process on a single computing node; (d) Distributed training process.}
	\label{fig:background_hgnn_training}
\end{figure*}

Fig.~\ref{fig:background_hgnn_training} depicts the training process of the HGNN models. The process of a single training epoch can generally be divided into four main steps as illustrated in Fig.~\ref{fig:background_hgnn_training}(c). Firstly, during forward propagation, the embeddings of target vertices are computed according to the procedural steps as in the model formula. Secondly, the loss computation (LC) stage transforms these embeddings into the vector space of classification categories, subsequently generating a probability distribution used to compute the loss function. Thirdly, backward propagation calculates the gradient of each model parameter relative to the loss function, employing the chain rule to determine the direction for parameter adjustments that yield the most rapid loss reduction. Finally, in the parameters update (PU) stage, model parameters are adjusted based on these gradients and a predefined learning rate.

\subsubsection{Full-batch and Mini-batch Training}

From the perspective of training methods, HGNN training can be categorized into two primary approaches: full-batch training, where the entire graph dataset is processed per epoch; and mini-batch training, which entails iterative processing of multiple mini-batches consisting of sets of target vertices and their respective neighbors within each epoch. As shown in Fig.~\ref{fig:background_hgnn_training}(a), semantic graphs are first constructed from the original HetG based on the relation type or predefined metapaths when performing full-batch training. Subsequently, multiple epochs are executed to iteratively update the parameters as illustrated in Fig.~\ref{fig:background_hgnn_training}(c). 

Mini-batch training differs in that the input data during execution does not encompass the entire graph structure. Instead, target vertices are partitioned into multiple groups, and neighbors of these vertices are sampled based on specified traversal depth and the number of sampled neighbors, forming individual mini-batches as depicted in Fig.~\ref{fig:background_hgnn_training}(b). Notably, when employing the mini-batch training method, the necessity of performing the SGB stage to construct a complete semantic graph is obviated. Instead, direct sampling is conducted, which can be regarded as a subset of the SGB execution process. Each training epoch involves the independent execution of multiple batches, with gradient updates performed autonomously for each batch, thus reducing resource requirements and improving the model's ability to handle large-scale datasets.

\subsubsection{Single-node and Distributed Training}

Distributed training has emerged as a solution to the limitations associated with the memory and computational capacity of a single machine. In practical applications, distributed training commonly employs mini-batch methods to achieve rapid convergence while preserving model accuracy~\cite{most_distributed_training}. As depicted in Fig.~\ref{fig:background_hgnn_training}(d), each computing node in distributed training independently processes one batch. Following backward propagation to obtain gradients, synchronization across nodes is essential through an \textit{All-reduce} operation, which ensures consistency of model parameters across all nodes at the outset of each epoch.

The principal benefits of distributed training encompass enhanced performance, facilitating parallel processing across multiple nodes; and expanded model scalability, enabling the training of more intricate models and processing of large-scale datasets. Moreover, this approach provides fault tolerance, as the failure of a single node does not necessarily disrupt the entire training process.

\subsubsection{Workload Distribution}

As shown in Fig.~\ref{fig:background_hgnn_training}(a) and Fig.~\ref{fig:background_hgnn_training}(b), the processes of SGB and mini-batch sampling are integral components of input data preparation, typically classified as data preprocessing. 
In contemporary mainstream training platforms, a heterogeneous configuration is frequently adopted, comprising a central host CPU and peripheral computing nodes, typically GPUs. Preprocessing tasks are typically executed by the CPU, which subsequently transfers the prepared data to the computing nodes for execution. These nodes are tasked with executing the entire training iterations and producing the final well-optimized model parameters.

\section{Characterization Methodology}
\label{sec:methodology}
To comprehensively and thoroughly characterize the training process of HGNNs, we adopt a rational and rigorous evaluation approach. In this section, we initially outline our experimental setup, encompassing the experimental platform, software framework, model and dataset selection. Then we delineate our evaluation methods.

\subsection{Experimental Setup}

\subsubsection{Platforms}

\begin{wraptable}{r}{0.5\textwidth}
\vspace{-12pt}
\centering
\caption{Platform Configurations}
\label{tab:platform}
\renewcommand\arraystretch{1.2}
\resizebox{0.5\textwidth}{!}{
    \begin{tabular}{|c|l|}
    \hline
                           & Configuration                       \\ \hline
    CPU                    & Intel(R) Xeon(R) Platinum 8350C CPU \\ \hline
    GPU                    & NVIDIA A100 80GB SXM GPU            \\ \hline
    OS                     & Ubuntu 20.04.5 LTS                  \\ \hline
    Framework              & Deep Graph Library 1.0.2            \\ \hline
    \multirow{2}{*}{Tools} & Nsight Compute 2021.2.1.0           \\ \cline{2-2} 
                           & Nsight Systems 2023.4.1.97          \\ \hline
    \end{tabular}
}
\scriptsize
\vspace{-12pt}
\end{wraptable}

The configuration of our experimental platform is detailed in Table~\ref{tab:platform}, featuring computing nodes equipped with four advanced A100 GPUs. The four GPUs are organized into two groups, with intra-group connections facilitated by NVLink buses, while inter-group connections are established through PCIe buses. We utilize Nsight Systems and Nsight Compute tools to capture detailed performance metrics data of execution. Regarding the software framework, we choose DGL~\cite{DGL}, which emerges as one of the most prominent GNN frameworks, typically outperforming PyG~\cite{PyG} in terms of runtime efficiency and energy consumption~\cite{characterize_PyG_DGL}. All the experiments are conducted utilizing 32-bit floating-point data format.

\subsubsection{HGNN Models}

We conduct experiments on three mainstream HGNN models, namely HAN~\cite{HAN}, RGCN~\cite{R-GCN} and RGAT~\cite{R-GAT} as shown in Table~\ref{tab:hgnn_info}, each with its own representative features. Specifically, RGCN first extends the conventional GCN~\cite{GCN} to handle HetGs by applying separate GCN convolutions to each semantic graph and aggregating the results through summation (SUM). RGAT builds upon this by introducing attention mechanisms in NA stage, with semantic fusion performed via SUM as well. HAN further enhances this by introducing attention mechanisms in SF stage, enabling the model to focus on important semantic graphs. In summary, the selected models are representative and widely adopted within the field, offering a comprehensive reflection of HGNN models as a whole.

\begin{table}[!t]
\centering
\caption{Information of HGNN models.}
\label{tab:hgnn_info}
\renewcommand\arraystretch{1.2}
\setlength\tabcolsep{2pt}%
\resizebox{1.0\textwidth}{!}{
\begin{tabular}{|c|c|c|c|c|c|c|c|}
\hline
Model & \#Layers & \#Hidden Dimension & \#Attention Heads & SGB      & FP                    & NA            & SF            \\ \hline
HAN   & 1       & 64                 & 8                & Metapath & Linear Transformation & Attention SUM & Attention SUM \\ \hline
RGCN  & 3       & 64                 & \diagbox{}{}                & Relation & Linear Transformation & MEAN          & SUM           \\ \hline
RGAT  & 3       & 64                 & 8                & Relation & Linear Transformation & Attention SUM & SUM          \\ \hline
\end{tabular}
}
\scriptsize
\end{table}

\subsubsection{Benchmark Datasets}
We employ four widely-used HetG datasets as benchmark datasets: ACM, IMDB, DBLP, and OGBN-MAG (MAG), as detailed in Table~\ref{tab:datasets}. ACM, DBLP, and MAG are citation datasets focused on academic publications, capturing intricate relationships between papers, authors, and institutions, making them well-suited for tasks such as citation analysis and academic recommendation systems. IMDB is a movie dataset containing detailed information about films, actors, and directors, commonly employed in movie recommendation systems and social network analysis. Ranging in size from tens of thousands to billions of edges, they offer comprehensive graph representations across a wide range of real-world applications.

\begin{table}[!ht]
\centering
\caption{Information of HetG datasets.} 
\label{tab:datasets}
\vspace{-4pt}
\renewcommand\arraystretch{1.2}
\setlength\tabcolsep{2pt}%
\resizebox{0.8\textwidth}{!}{
\begin{tabular}{|c|c|c|c|c|}
\hline
\textbf{Dataset} &
  \textbf{\#Vertex} &
  \textbf{\#Feature} &
  \textbf{\#Edge of Relations} &
  \textbf{\#Edge of Metapaths} \\ \hline
\multirow{4}{*}{ACM} &
  paper (P): 3025 &
  P:1902 &
  \multirow{4}{*}{\begin{tabular}[c]{@{}c@{}}AP: 9936 PA: 9936\\ PS: 3025 SP: 3025\end{tabular}} &
  \multirow{4}{*}{\begin{tabular}[c]{@{}c@{}}PAP: 29436\\ PSP: 2200581\\ PAPSP: 3666289\end{tabular}} \\ \cline{2-3}
 &
  author (A): 5912 &
  A: 1902 &
   &
   \\ \cline{2-3}
 &
  subject (S): 56 &
  S: 1902 &
   &
   \\ \cline{2-3}
 &
  term (T):1902 &
  T: - &
   &
   \\ \hline
\multirow{4}{*}{IMDB} &
  movie (M): 4278 &
  M: 3066 &
  \multirow{4}{*}{\begin{tabular}[c]{@{}c@{}}AM: 12828 DM: 4278\\ MA: 12828 MD: 4278\end{tabular}} &
  \multirow{4}{*}{\begin{tabular}[c]{@{}c@{}}MDM: 17446\\ MAM: 85358\\ MAMDM: 338517\end{tabular}} \\ \cline{2-3}
 &
  director (D):2081 &
  D: 2081 &
   &
   \\ \cline{2-3}
 &
  actor(A):5257 &
  A: 5257 &
   &
   \\ \cline{2-3}
 &
  keyword (K):7971 &
  K: - &
   &
   \\ \hline
\multirow{4}{*}{DBLP} &
  author (A): 4057 &
  A: 334 &
  \multirow{4}{*}{\begin{tabular}[c]{@{}c@{}}AP: 19645 PA: 19645\\ VP: 14328 PV: 14328\\ TP: 85810 PT: 85810\end{tabular}} &
  \multirow{4}{*}{\begin{tabular}[c]{@{}c@{}}APA: 11113\\ APVPA: 5000495\\ APTPA: 7043571\end{tabular}} \\ \cline{2-3}
 &
  paper (P): 14328 &
  P:14328 &
   &
   \\ \cline{2-3}
 &
  term (T):7723 &
  T: 7723 &
   &
   \\ \cline{2-3}
 &
  venue(V):20 &
  V: 20 &
   &
   \\ \hline
\multirow{4}{*}{\begin{tabular}[c]{@{}c@{}}OGBN-MAG\\ (MAG)\end{tabular}} &
  author (A):1134649 &
  A: 129 &
  \multirow{4}{*}{\begin{tabular}[c]{@{}c@{}}AI: 1043998 IA: 1043998\\ AP: 7145660 PA: 7145660\\ PF: 7505078 FP: 7505078\end{tabular}} &
  \multirow{4}{*}{\begin{tabular}[c]{@{}c@{}}PAP: 65933339\\ PPAP: 614471897\end{tabular}} \\ \cline{2-3}
 &
  paper (P): 736389 &
  P: 14328 &
   &
   \\ \cline{2-3}
 &
  feld (F): 59965 &
  T: 7723 &
   &
   \\ \cline{2-3}
 &
  institute (l):8740 &
  V: 20 &
   &
   \\ \hline
\end{tabular}
}
\vspace{-4pt}
\scriptsize
\end{table}

\subsection{Evaluation Methods}

A fundamental and critical step in profiling involves defining the scope of the profiling process. We utilize the CUDA interface provided by PyTorch to initiate and terminate the \textit{cudaProfiler}. Furthermore, \textit{NVTX} tags are employed to demarcate the code regions designated for profiling, thereby distinguishing between different execution phases.

Nsight Systems is employed to deliver a comprehensive analysis of system-level performance, offering insights into the complete application execution process, encompassing both GPU and CPU activities, as well as their interactions with system resources. In contrast, Nsight Compute is dedicated to providing an in-depth analysis of CUDA kernel executions on the GPU. It supplies detailed performance metrics at the GPU instruction level, including execution time, memory access patterns, and instruction efficiency for each CUDA kernel function.

Due to the GPU initialization processes involved in the initial training epochs, unless otherwise specified, the data presented in this paper are the geometric mean (GM) of results from the 5 epochs after excluding the first 3 epochs, which ensures the accuracy and reliability of the obtained results.

\section{Single-GPU Training}
\label{sec:single}

Understanding the execution process on a single GPU is essential to comprehending the execution behavior and characteristics of HGNNs. In this section, we conduct a detailed performance analysis of two principal training methods, full-batch and mini-batch training, executed on a single GPU to identify the execution bottlenecks of HGNNs under different training methods.

\subsection{Full-batch Training}
\label{sec:full-batch}
Full-batch training enables the model to access the entire graph for thorough parameter updates, leading to accurate and stable gradient estimates; however, it necessitates substantial memory resources. In this section, we provide a detailed quantitative analysis of execution time, execution bounds, memory access patterns, and instruction issue stalls during full-batch training. Furthermore, we investigate the impact of variations in metapath properties on execution performance. Due to an Out of Memory (OOM) issue encountered while conducting full-batch training on the MAG dataset with the A100 GPU, this section includes results solely for the three smaller datasets. Furthermore, as the SGB stage is executed only once throughout the entire training process, its associated analysis will be presented in Section~\ref{sec:SGB}.

\subsubsection{Execution Time Analysis}
\label{sec:single_gpu_fullbatch_time_analysis}

In this section, we provide a comprehensive analysis of the execution time during full-batch HGNN training. Our analysis focuses on identifying the principal execution components from two perspectives encompassing various execution stages and the distinct kernels within each stage.


\begin{figure*}[!ht] 
	\centering
	\includegraphics[width=1\textwidth]{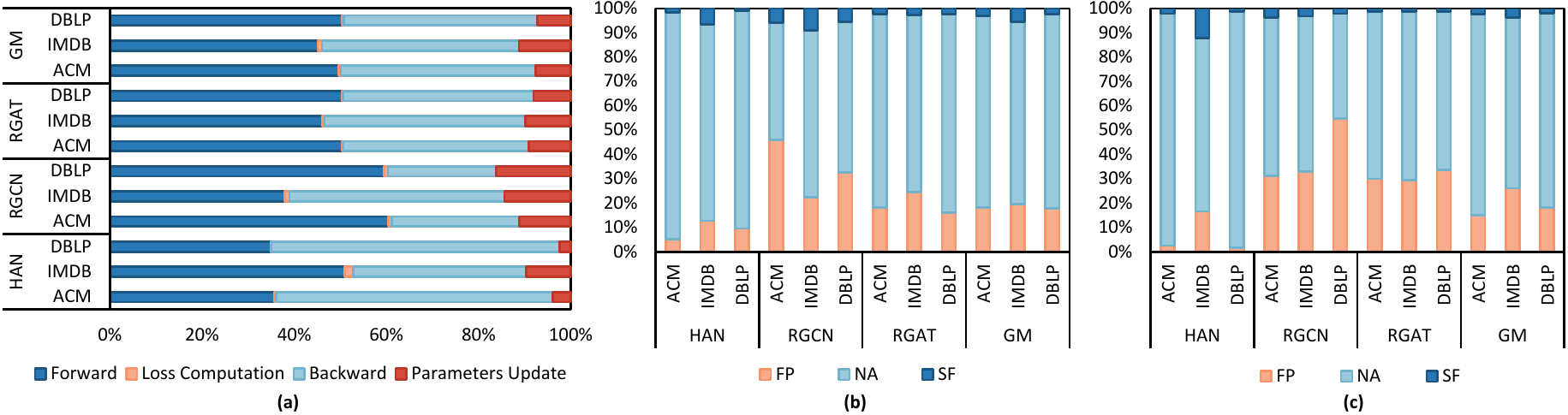}
	\caption{Time breakdown of HGNN training by phase: (a) The whole training process; (b) Forward; (c) Backward.}
	\label{fig:single_gpu_time_breakdown_by_phase}
\end{figure*}

\textbf{Time Breakdown by Stage.} Breaking down the execution time by stages allows us to understand the time proportion of each execution stage, thereby identifying the main execution stages for targeted optimization. Fig.~\ref{fig:single_gpu_time_breakdown_by_phase} shows the profiling results of the time breakdown by stages. From the overall training perspective, \marknumber{1} \textit{forward propagation is more time-consuming than backward propagation}. This phenomenon occurs because, although the backward stage involves the reverse operations of forward propagation, it can directly reuse numerous intermediate results from the forward process. The experimental result in Fig.~\ref{fig:single_gpu_time_breakdown_by_phase}(a) indicates that forward propagation accounts for an average of 48.33\% of the total execution time across different HGNN models and datasets, while backward propagation averages 42.37\%. In terms of HGNN execution stages, \marknumber{2} \textit{the NA stage dominates the most execution time both in forward and backward propagation, which is considered the predominant stage across all the HGNN stages}. This is due to the fact that each edge in every semantic graph necessitates a distinct execution process during the NA stage, making the load during NA stage the heaviest. As shown in Fig.~\ref{fig:single_gpu_time_breakdown_by_phase}(b) and Fig.~\ref{fig:single_gpu_time_breakdown_by_phase}(c), the NA stage accounts for an average of 77.80\% of the execution time in the forward propagation and 77.53\% in the backward propagation.

\textbf{Time Breakdown by Kernel.} Analyzing the execution time distribution of CUDA kernels offers valuable insights into the primary kernels utilized during each stage of execution, thereby elucidating their core execution characteristics. We categorize these kernels into four distinct groups based on their specific computational tasks: dense-dense matrix multiplication (DeMM) kernel (DM-Type), topology-based matrix operation kernel (TB-Type), element-wise computation kernel (EW-Type), and data rearrangement kernel (DR-Type), as detailed in our prior work~\cite{understand_HGNN}.

The DM-Type kernels perform DeMM tasks, such as \textit{sgemm}, and typically exhibit a regular execution pattern with a high compute-to-memory-access ratio. The TB-Type kernels manage computational operations based on the irregular topologies of graphs, exemplified by operations like \textit{SpMMCsr} and \textit{SDDMMCoo} (sampled dense-dense matrix multiplication, \textit{SDDMM}), and often demonstrate an irregular execution pattern due to the uneven neighbor connection patterns in graphs. The EW-Type kernels execute element-wise computational operations on sets of vectors or matrices, represented by \textit{elementwise\_kernel} (\textit{EleWise}), \textit{matrix\_scalar\_kernel} (\textit{MatScla}), and \textit{reduce\_kernel} (\textit{Reduce}), typically characterized by a low compute-to-memory-access ratio. The DR-Type kernels are specialized in data rearrangement tasks on matrices, such as \textit{CatArrayBatchedCopy} (\textit{Concat}) and \textit{DeviceRadixSortSingleTileKernel} (\textit{Sort}), and involve a significant amount of data movement operations.

\begin{figure*}[!ht] 
	\centering
	\includegraphics[width=1\textwidth]{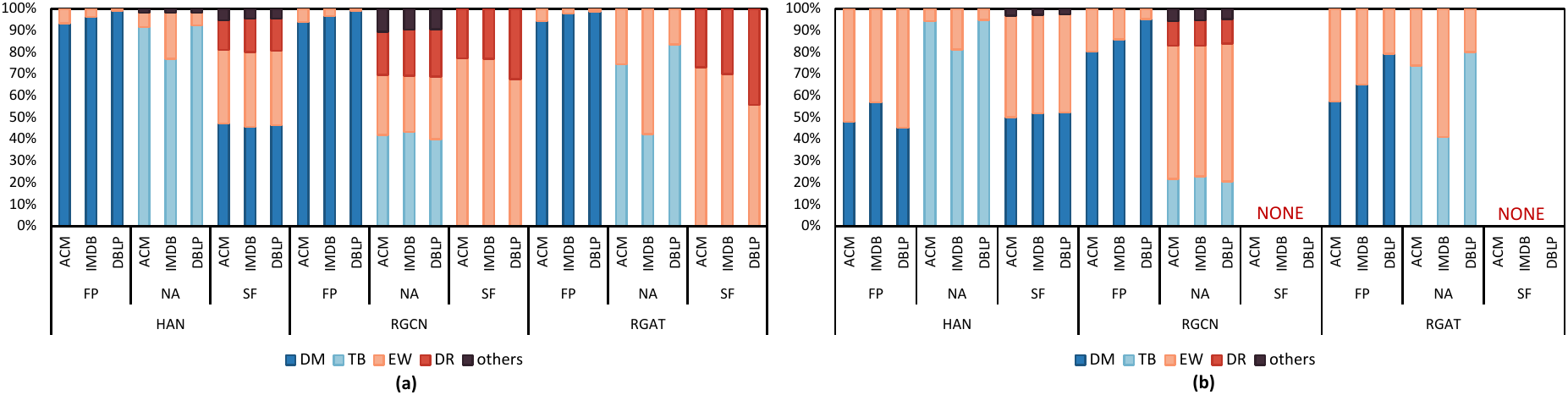}
	\caption{Time breakdown of HGNN training by kernel: (a) Forward; (b) Backward ("NONE" indicates that there are no CUDA kernels invoked here).}
	\label{fig:single_gpu_time_breakdown_by_kernel}
\end{figure*}

From an overall perspective, \marknumber{3} \textit{the distribution of primary executing kernel types during the forward and backward propagation of the same stage exhibits significant similarity, whereas the proportion of time consumed by these kernel types varies between the forward and backward propagation}. The resemblance in the distribution of kernel types stems from backward propagation being the inverse operation of forward propagation in model computations. The disparities in the distribution of kernel execution times originate from discrepancy in input data and computational loads during backward propagation. For example, as shown in Fig.~\ref{fig:single_gpu_time_breakdown_by_kernel}(a), in the forward propagation, the FP stage is predominantly occupied by the DM-Type kernel (mainly \textit{sgemm}), taking up an average of 96.48\% of the total kernel execution time across various models and datasets. The EW-Type kernels (mainly \textit{EleWise}) only occupy a small fraction which is less than 4\%. However, during the backward propagation, the proportion of time spent on EW-Type kernels notably escalates, averaging 25.78\% of the total kernel execution duration. The underlying cause of this phenomenon lies in the vertex-type-specific nature of the weight matrix employed during the FP stage which will be further explained in Section~\ref{sec:gnn_comparison_seprate_FP}.

Note that the RGCN and RGAT models do not invoke any CUDA kernels during the backward propagation of SF stage as in Fig.~\ref{fig:single_gpu_time_breakdown_by_kernel}(b). This is due to their adoption of a direct summation aggregation approach in the forward propagation of SF stage. Moreover, in a broader context, \marknumber{4} \textit{addition operations during forward propagation necessitate no computational effort during the backward pass, and this phenomenon is observed in every execution stage}. For instance, during the projection process in the FP stage, linear transformations typically involve the addition of a bias term, applied to the resulting matrix through element-wise addition. Additionally, in the NA stage, a significant amount of element-wise vector addition is used to aggregate the feature vectors of neighboring vertices, with a similar operation occurring in the SF stage. Consider matrices, vectors of the same dimensions, or scalars denoted as $A$, $B$, and $C$, with the relation $A = B + C$. If the forward propagation loss is defined as $Loss = f(A) = f(B + C)$, given the gradient of $A$ with respect to $Loss$: $\frac{\partial L}{\partial A}$, according to the chain rule of gradient propagation, since the partial derivatives of $A$ with respect to $B$ and $C$ are both 1, the gradients of $B$ and $C$ can be directly obtained from $\frac{\partial L}{\partial A}$. Specifically, the gradients of $B$ and $C$ are equal to the gradient of $A$, without any additional computation. This rule can be formally represented by the equation: $Loss = f(A) = f(B + C)\ \rightarrow \frac{\partial L}{\partial B} = \frac{\partial L}{\partial C} = \frac{\partial L}{\partial A}$.



\subsubsection{Execution Bounds Analysis}
\label{sec:single_gpu_fullbatch_execution_bound_analysis}

In this section, we perform a comprehensive analysis of the performance metrics of CUDA kernels predominantly invoked at each execution stage, with the aim of identifying the hardware resource bounds encountered throughout the various stages.

\begin{figure*}[!htbp] 
	\centering
	\vspace{-12pt}
	\includegraphics[width=0.96\textwidth]{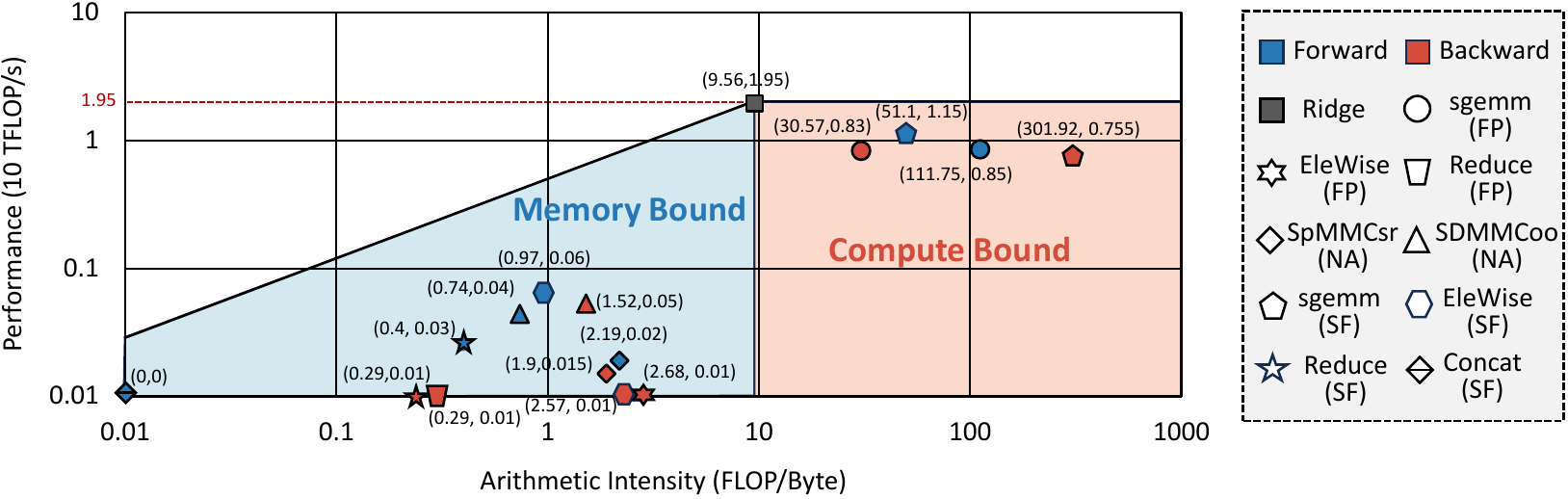}
	\caption{The roofline model for kernels under single-precision floating-point operations.}
	\label{fig:single_gpu_roofline}
\end{figure*}

\textbf{Forward Propagation Analysis.} The forward propagation process constitutes the primary execution phase of HGNN model training. Characterizing and analyzing the execution bounds during this phase is essential for understanding the requirements of various hardware resources throughout the model's execution. Table~\ref{tab:kernel_profiling_details} presents the performance metrics of the primary kernels at each execution stage during the training of the HAN model on the DBLP dataset. Based on the data in the table, we formulate the \textit{Roofline} model~\cite{roofline} shown in Fig.~\ref{fig:single_gpu_roofline}, which elucidates the execution bounds of each stage encountered during both forward and backward propagation.

\marknumber{5} \textit{During the forward propagation, a distinct hybrid execution pattern emerges between stages, each characterized by unique execution bounds, resulting in varying demands on hardware resources.} This is due to the distinct execution behaviors exhibited by different stages of the HGNN models. To be specific, for the FP stage, the kernels that occupy the majority of execution during forward propagtion is mainly \textit{sgemm}, which primarily performs DeMM operations. As shown in Table~\ref{tab:kernel_profiling_details}, due to its regular memory access pattern, it attains an L2 cache hit rate of 86.14\%, while its high arithmetic density of 111.75 FLOP/Byte indicates a substantial computational demand. In Fig.~\ref{fig:single_gpu_roofline}, the \textit{sgemm} kernel during the forward propagation is situated in the compute-bound region. The NA stage primarily involves graph-topology-based and element-wise operations, mainly invoking \textit{SpMMCsr} and \textit{SDDMMCoo} kernels as presented in Table~\ref{tab:kernel_profiling_details}. Taking the former as an example, its DRAM bandwidth utilization is 53.11\%, but its arithmetic density is only 2.19 FLOP/Byte, indicating its high demand for memory resources. In Fig.~\ref{fig:single_gpu_roofline}, it lies within the memory-bound region. In the SF stage, the \textit{sgemm} kernel is initially utilized for computing attention weights, subsequent to which \textit{EleWise} and \textit{Reduce} kernels are employed for aggregating features from diverse semantic graphs. According to the aforementioned analysis, this stage demonstrates an initial manifestation of compute-bound behavior, succeeded by memory-bound characteristic as depicted in Fig.~\ref{fig:single_gpu_roofline}.

\begin{table*}[!thbp]
    \caption{Profiling results of major CUDA kernels on HAN model with DBLP dataset.} \label{tab:kernel_profiling_details}
    \centering
    \setlength\tabcolsep{5pt}%
	\renewcommand\arraystretch{0.85}
    \resizebox{0.99\textwidth}{!}{
\centering
    \begin{tabular}{ccccccccc}
    \toprule
      \begin{tabular}[c]{@{}c@{}} Stage  \end{tabular}  &
      \begin{tabular}[c]{@{}c@{}} Kernel \\ Name  \end{tabular}  & 
      \begin{tabular}[c]{@{}c@{}} Kernel \\ Type  \end{tabular}  & 
      \begin{tabular}[c]{@{}c@{}} Time   \\ (\%) \end{tabular}     & 
      \begin{tabular}[c]{@{}c@{}} Achieved Peak \\ Performance (\%) \end{tabular} & 
      \begin{tabular}[c]{@{}c@{}} DRAM BW \\ Utilization (\%) \end{tabular}  &
      \begin{tabular}[c]{@{}c@{}} Shared Memory BW \\ Utilization (\%) \end{tabular}  &
    \begin{tabular}[c]{@{}c@{}} L2 Cache \\ Hit  Rate (\%) \end{tabular} &
    \begin{tabular}[c]{@{}c@{}} Arithmetic \\Intensity (FLOP/Byte) \end{tabular}\\ \midrule \midrule
     \multicolumn{9}{c}{ \textbf{Feature Projection} }    \\ \midrule \midrule
      Forward  & sgemm  & DM  & 98.94\% & 56.15\% & 10.15\% & 38.17\% & \textbf{86.14}\%  & \textbf{111.75}  \\ \midrule
        & sgemm &DM & 45.30\% & 57.26\% & 14.14\% & 37.16\% & 62.43\%  & 30.57  \\
      Backward  & \textbf{EleWise} &EW & 36.76\% & 0.51\% & 15.88\% & 0.80\% & 76.25\%  & 2.68  \\
        & \textbf{Reduce} &EW & 17.94\% & 0.50\% & 21.14\% & 0.72\% & 29.91\%  & 0.29  \\ \midrule \midrule
        
    \multicolumn{9}{c}{\textbf{Neighbor Aggregation} }       \\ \midrule \midrule
       & SpMMCsr &TB & \textbf{84.16}\% & 1.00\%  & \textbf{53.11}\%  & 0.23\% &63.19\%  & \textbf{2.19}\\
      Forward & SDDMMCoo &TB & \textbf{8.21}\% & 2.32\%  & 32.10\%  &1.78\% & 71.17\%  & 0.74\\
       & EleWise  &EW & 5.42\%   & 0.89\%  & 31.60\% &1.30\% & 86.54\%  & 0.21\\ \midrule
       & SpMMCsr &TB & \textbf{48.58}\% & 7.90\%  & 5.79\%  &0.18\% & 59.73\%  & 1.90\\
       Backward & SDDMMCoo &TB & \textbf{46.21}\% & 2.79\%  & 27.92\%  &10.96\% & 70.13\%  & 1.52\\
        & EleWise  &EW & 5.04\%   & 0.63\%  & 30.32\% &0.95\% & 75.11\%  & 1.09\\ \midrule \midrule
        
        \multicolumn{9}{c}{\textbf{Semantic Fusion} }        \\ \midrule \midrule 
        & sgemm  &DM & 46.43\% & 73.64\% & 11.12\% &24.35\% & 78.94\%  & 51.10\\
      Forward & EleWise  &EW & 21.02\%   & 3.2\%  & 26.97\% &1.78\% & 75.03\%  & 0.97\\
        & Concat  &DR  & 14.87\% & 0.00\%    & 45.78\% &0.23\% & 62.96\% & 0.00\\
        & Reduce &EW  & 13.16\% & 1.31\%    & 29.14\% &1.29\% & 54.00\% & 0.40\\ \midrule
    
        & sgemm  &DM & 52.22\% & 61.41\% & 4.33\% &39.99\% & 91.79\%  & 301.92\\
     Backward  & EleWise  &EW & 28.72\%   & 0.53\%  & 21.07\% &0.63\% & 82.31\%  & 2.57\\
        & Reduce &EW  & 16.39\% & 0.47\%    & 14.13\% &0.94\% & 47.32\% & 0.29\\

        \bottomrule
    \end{tabular}
}
\end{table*}

\textbf{Comparison of Backward and Forward Propagation.} As a crucial component of the training process, backward propagation shares some similarities with but also exhibits significant differences from forward propagation. Contrasting them aids in a deeper understanding of the varying hardware resource demands during HGNN training. Overall, \marknumber{6} \textit{the execution bounds in the backward and forward propagation exhibit similarity, yet the hybrid execution pattern in backward propagation is notably more intricate}. This phenomenon arises because, while backward propagation reverses the process of forward propagation, the application of the chain rule in gradient propagation may introduce operations within the same stage that deviate from those encountered in forward propagation. As shown in Table~\ref{tab:kernel_profiling_details}, in FP-Backward compared to FP-Forward stage, the proportion of time spent on \textit{sgemm} decreases from 98.94\% to 45.40\%, while the time occupied by \textit{EleWise} kernels facing memory-bound notably increases. In both forward and backward propagation, the NA stage primarily exhibits memory-bound. However, what differs is that in the NA-Backward stage, the proportion of time taken by the \textit{SDDMMCoo} kernel increases from 8.21\% to 46.21\%, which indicates the matrices involved in matrix multiplication during the NA stage of backward propagation are denser compared to those in forward propagation. This is due to the fact that, during backward propagation of NA stage, in addition to gradient propagation through edges, it is necessary to compute gradients of the vertex features first, which involves operating on dense matrices. The execution bounds exhibited by the SF stage in both backward and forward propagation are fundamentally similar.

\subsubsection{Memory Pattern Analysis}
\label{sec:single_gpu_fullbatch_memory_related}
In this section, we focus on the memory aspects, encompassing DRAM access, memory footprint, and cache hit rates, to elucidate the memory characteristics of various execution stages during the HGNN training process.

\begin{figure*}[!ht] 
	\centering
	\includegraphics[width=0.96\textwidth]{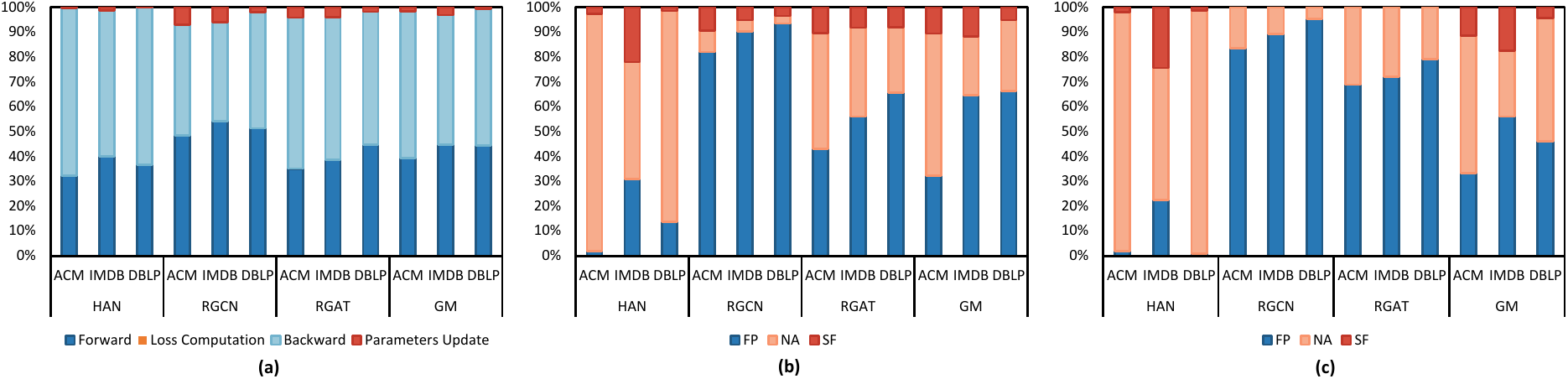}
	\caption{Breakdown of DRAM access volume of HGNN training by phase: (a) Training; (b) Forward; (c) Backward.}
	\label{fig:single_gpu_dram_access}
\end{figure*}

\textit{\textbf{DRAM Access Breakdown.}} Memory access latency is a critical factor contributing to performance degradation during model execution, and it is associated with high energy consumption. A comprehensive understanding of memory access patterns across various stages of HGNN training facilitates both performance and energy efficiency improvement. Fig.~\ref{fig:single_gpu_dram_access} illustrates the DRAM access volume breakdown of various models during the training process of HGNN on different datasets. From a training perspective, \marknumber{7} \textit{the backward propagation generally incurs more memory access compared to the forward propagation}. This phenomenon arises due to the reason that during the backward propagation, the model must access gradient data that matches the size of the dataset accessed in forward propagation. Additionally, the computation of gradients requires the accesses of numerous intermediate results preserved from the forward propagation. As presented in Fig.~\ref{fig:single_gpu_dram_access} (a), the backward propagation accounts for 55.38\% of the total memory access during the entire training process, while the forward propagation averages only 42.87\%. The discrepancy in DRAM access volume between backward and forward propagation varies across different models and datasets, primarily due to the differing quantities of intermediate results stored during forward propagation, which are influenced by the complexity of model structure. This discrepancy is most pronounced in the HAN model on the ACM dataset, where the proportion of DRAM access for backward propagation is 2.32$\times$ greater than that for forward propagation.

Looking specifically at the internal stages of the forward and backward propagation, there are significant differences in memory access distribution among different models. However, from the perspective of the same execution workload (same model and dataset), \marknumber{8} \textit{the stages dominating memory accesses during both backward and forward propagation remain consistent, with the proportions of memory accesses in corresponding stages between forward and backward propagation exhibiting notable similarity}. This phenomenon still stems from the fact that backward propagation largely mirrors the computation flow of forward propagation in reverse for most operations. Specifically, as shown in Fig.~\ref{fig:single_gpu_dram_access} (b) and Fig.~\ref{fig:single_gpu_dram_access}(c), in both forward and backward propagation, the HAN model primarily incurs memory accesses during the NA stage across various datasets, whereas for the RGCN and RGAT models, the primary memory accesses originate from the FP stage.

\begin{figure*}[!ht] 
	\centering
	\includegraphics[width=0.96\textwidth]{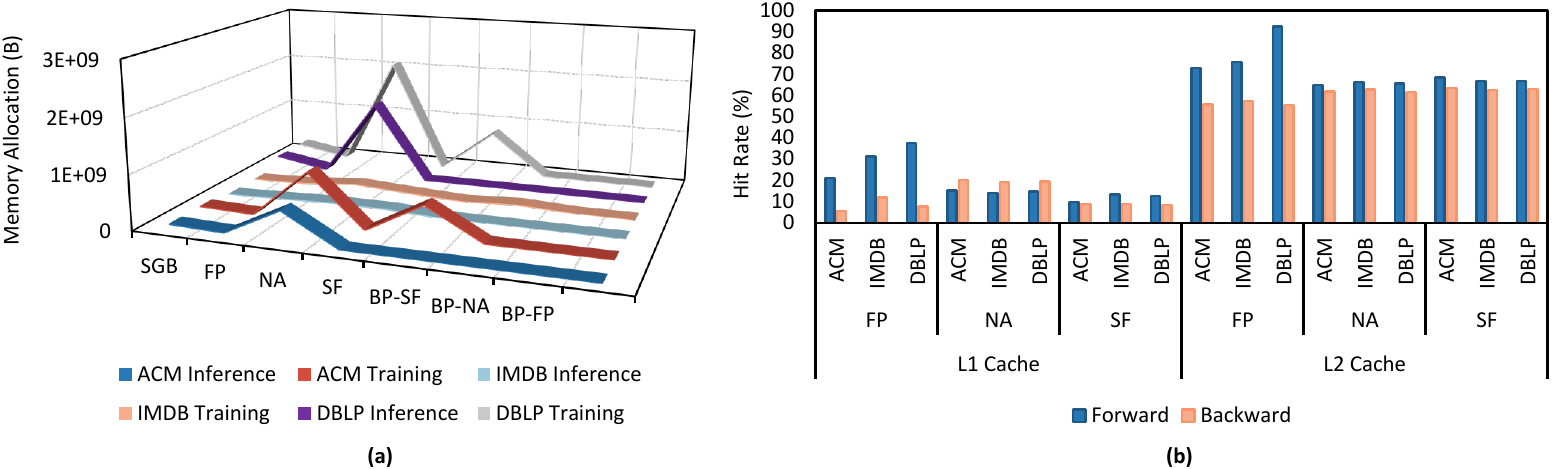}
	\caption{Memory-related profiling results of HGNN Training: (a) Memory Footprint; (b) Cache Hit Rate.}
	\label{fig:single_gpu_memory_related}
\end{figure*}

\textit{\textbf{Memory Footprint Analysis.}} As the scale of real-world graph data continues to grow, memory utilization emerges as a pivotal factor influencing the scalability of models applied to large-scale datasets. Conducting an analysis of the memory footprint facilitates the optimization of model scalability. Fig.~\ref{fig:single_gpu_memory_related}(a) illustrates the memory allocation in each execution stage during the training process of HGNN along the timeline. For the sake of convenience, only the specific cases of the HAN model on three different datasets are shown here, while the situations for the other two models are similar. Here, we compare the situations between the pure inference process and the complete training process to highlight the more urgent memory demands during the training process.

\marknumber{9} \textit{In comparison to pure inference, the training process necessitates a greater memory footprint, with the majority of memory allocation lying on the NA stage}. This distinction arises from inference's singular execution of forward propagation, which does not involve gradient computation and thereby obviates the storage of intermediate computational results essential for training during the forward propagation. Furthermore, the inference process does not require backward propagation, thus eliminating the associated memory allocation needed during training. And given that the numerous intermediate computation results requiring storage are associated with edges, the NA stage necessitates the most substantial memory allocation. Considering the case of DBLP dataset in Fig.~\ref{fig:single_gpu_memory_related}(a), in comparison to the inference process, the total memory allocation during training increases by 1.92$\times$, with the NA stage constituting 66.63\% of the total memory consumption.

\textit{\textbf{Cache Hit Rate Analysis.}} The cache hit rate effectively reflects data locality, thereby serving as a starting point for optimizing models from this perspective. Fig.~\ref{fig:single_gpu_memory_related}(b) illustrates the differences in cache hit rates for forward and backward propagation across various stages of execution on different models of the HAN model. As shown in the figure, in nearly all scenarios, \marknumber{10} \textit{the data locality during backward propagation tends to be lower compared to that observed during forward propagation}. This discrepancy primarily arises because backward propagation involves not only accessing graph data but also retrieving stored gradient information and various intermediate results, necessitating a broader range of memory accesses and longer reuse distance of data, leading to diminished data locality. For the FP, NA, and SF stages, the L2 cache hit rate during backward propagation is on average 23.71\%, 3.57\%, and 4.34\% lower than during forward propagation, respectively.

\subsubsection{Issue Stall Analysis} This section presents a comprehensive analysis of GPU instruction issue stalls observed throughout the training process. The emphasis is on elucidating both the temporal distribution and the underlying causes of these stalls at each stage of the process. Analyzing instruction issue stalls offers insights into operational bottlenecks from a distinct perspective, thereby facilitating the identification of targeted optimization strategies.

\begin{figure*}[!ht] 
	\centering
	\includegraphics[width=0.96\textwidth]{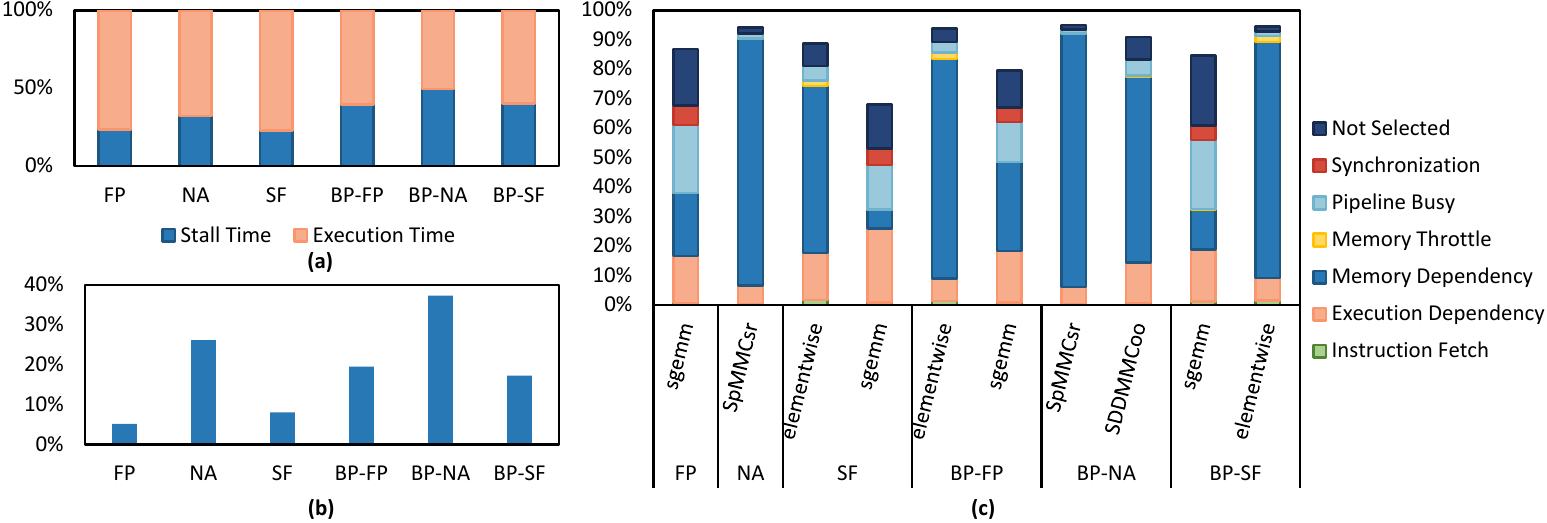}
	\caption{Issue stall of HAN model on DBLP dataset: (a) The ratio of stall time to elapsed time; (b) Memory dependency stall ratio to elapsed time; (c) Breakdown of issue stall reasons of different stages.}
	\label{fig:single_gpu_issue_stall}
\end{figure*}

\textit{\textbf{Overall Profiling Results.}} Fig.~\ref{fig:single_gpu_issue_stall}(a) presents the ratio of stall time to execution time for different execution stages of the training process. From a comprehensive viewpoint, the prevalence of instruction issue stall across different stages throughout the training process is notable and merits careful consideration, averaging 33.21\% across different stages. Fig.~\ref{fig:single_gpu_issue_stall}(c) illustrates the breakdown of instruction issue stall reasons for the main CUDA kernels invoked at each execution stage. It can be observed that \marknumber{11} \textit{the vast majority of kernel stalls originate from memory dependency except for \textit{sgemm} kernel, and the proportion of stall time attributable to memory dependency during the NA stage is the highest among all execution stages}. This is due to the fact that, during the training process of HGNNs, many kernels other than \textit{sgemm} involve extensive read and write operations on irregular graph data, which is particularly evident in the graph-structure-related kernels such as \textit{SpMMCsr} and \textit{SDDMMCoo} employed in NA stage. These access patterns are usually highly irregular, making it difficult for the accessed data to be fully cached. Furthermore, elementwise operations in HGNN training may require accessing data distributed across non-contiguous memory locations, resulting in frequent and unpredictable memory access patterns. Therefore, stalls caused by memory dependencies dominate. As shown in Fig.~\ref{fig:single_gpu_issue_stall}(c), for kernels other than \textit{sgemm}, memory dependency accounts for an average of 74.11\% of the instruction issue stalls. In contrast, for the \textit{sgemm} kernel, memory dependency accounts for an average of only 15.51\%, with the dominant stalls being caused by execution dependency, which averages 18.86\%. Fig.~\ref{fig:single_gpu_issue_stall}(b) illustrates that the NA stage exhibits the highest ratio of stalls caused by memory dependency relative to total elapsed time across all execution stages during both forward and backward propagation.

\textit{\textbf{Comparison of Backward and Forward.}} Fig.~\ref{fig:single_gpu_issue_stall}(b) presents the proportion of stalls attributed to memory dependency in relation to the total elapsed time for each execution stage of HGNN training during both forward and backward propagation. \marknumber{12} \textit{The fraction of overall execution time attributed to memory dependency during each stage of backward propagation surpasses that observed in forward propagation}. The heightened memory access volume during the backward propagation stage, coupled with increased data type diversity, leads to diminished data locality and reduced cache hit rates, as detailed in Section~\ref{sec:single_gpu_fullbatch_memory_related}. As shown in Fig.~\ref{fig:single_gpu_issue_stall}(b), the average proportion of memory dependency during each stage of backward propagation is 2.26$\times$ higher than that during forward propagation.


\subsubsection{Exploring Metapath Changes}
\label{sec:single_gpu_fullbatch_metapath_change}

In this section, we explore the performance impact on each execution stage as the length and number of metapaths vary. As the number of metapaths increases significantly, it can greatly expand the size of the graph data, potentially leading to an OOM issue on a single GPU. Therefore, in the experiments in this section, we use the IMDB dataset as a representative example, focusing on the HAN model. And the performance metrics for each stage covered in this section are obtained by averaging forward and backward propagation.

\begin{figure*}[!ht] 
	\centering
	\includegraphics[width=0.8\textwidth]{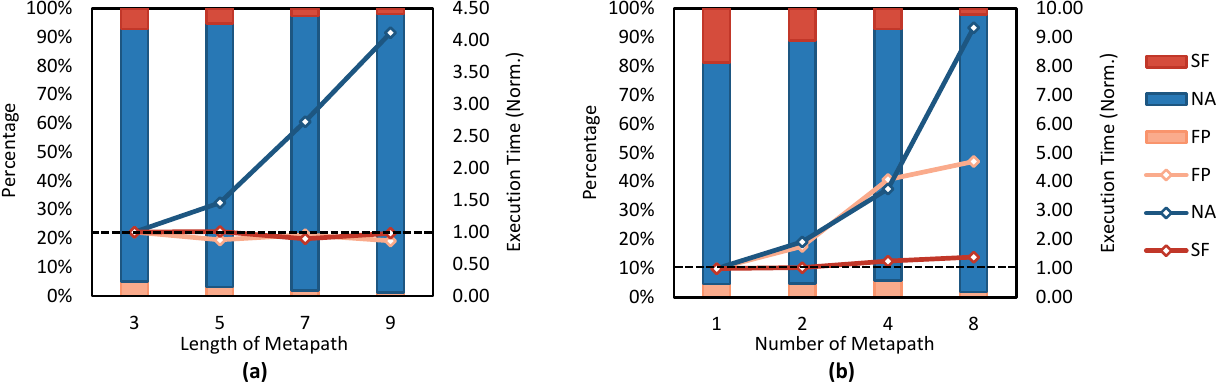}
	\caption{Execution time breakdown as metapath changes: (a) Length of metapath increases; (b) Number of metapath increases.}
	\label{fig:single_gpu_metapath_changes}
\end{figure*}

\textit{\textbf{Increase in Length of Metapaths.}} Exploring the performance changes at different stages due to variations in metapath length is crucial, as longer metapaths can assist models in capturing more complex relationship patterns and context information over greater distances. Fig.~\ref{fig:single_gpu_metapath_changes}(a) shows the execution time of different stages across varying metapath lengths. Obviously, \marknumber{13} \textit{increase in the length of the metapath only significantly increases the execution time of NA stage, while the FP and SF stages are almost unaffected.} The rationale behind this lies in the fact that the length increase of the metapath does not alter the vertex types or the number of semantic graphs, thereby maintaining the workloads on the FP and SF stages. However, elongating the metapath results in denser semantic graphs~\cite{understand_HGNN}, characterized by a higher volume of edges. This directly translates into increased execution times during the NA stage. As depicted in Fig.~\ref{fig:single_gpu_metapath_changes}(a), an increase in metapath length from 3 to 9 correlates with a 4.12$\times$ rise in execution time for the NA stage, while the FP and SF stages exhibit negligible change in execution times.


\textit{\textbf{Increase in Number of Metapaths.}} Exploring the impact of varying the number of metapaths on the execution performance of different stages is essential, as a greater number of metapaths can provide the model with richer semantic graph information, thereby enhancing model accuracy. Fig.~\ref{fig:single_gpu_metapath_changes}(b) illustrates the impact of increasing the number of metapaths while keeping the metapath length constant on the performance at various execution stages. According to the figure, \marknumber{14} \textit{increase in the number of metapaths leads to increased execution times across all stages of HGNN, with the NA stage showing the most noticeable growth trend}. As the number of metapaths increases, so does the number of semantic graphs, resulting in a higher volume of vertices and edges to process. Consequently, each execution stage experiences a proportional increase in workload. Notably, the NA stage shows the most pronounced growth, primarily because the rise in edges outpaces that of vertices and even semantic graphs. As illustrated in Fig.~\ref{fig:single_gpu_metapath_changes}(b), expanding the metapath count from 1 to 8 leads to execution time increases of 4.71$\times$ for FP, 9.34$\times$ for NA, and 1.40$\times$ for SF.

\subsection{Mini-batch Training}
\label{sec:mini-batch}


As the size of real-world graph datasets continues to expand, mini-batch training has increasingly become the predominant method for model training, which leads to faster convergence and improved generalization performance compared to full-batch training. Aside from the additional mini-batch sampling phase, the execution behavior and characteristics are identical to those in full-batch training. Consequently, this section primarily focuses on profiling the mini-batch sampling phase. Unless otherwise specified, the batch size used for sampling is 256 for ACM, IMDB and DBLP datasets. Due to the larger scale of the MAG dataset, its batch size is set to 1024. The sampling method employed is \textit{Random Walk Neighbor Sampler}~\cite{GraphSage}. Unless specified otherwise, the sampling process is executed on the CPU.

\subsubsection{Execution Analysis}
\label{sec:single_gpu_minibatch_execution_analysis}
This section provides a detailed analysis of the execution time breakdown during mini-batch training, compares the performance of GPU sampling with that of CPU sampling, and examines various performance metrics of the CUDA kernels employed in GPU sampling to elucidate the characteristics of the sampling operation.

\begin{figure*}[!ht] 
	\centering
	\includegraphics[width=1.0\textwidth]{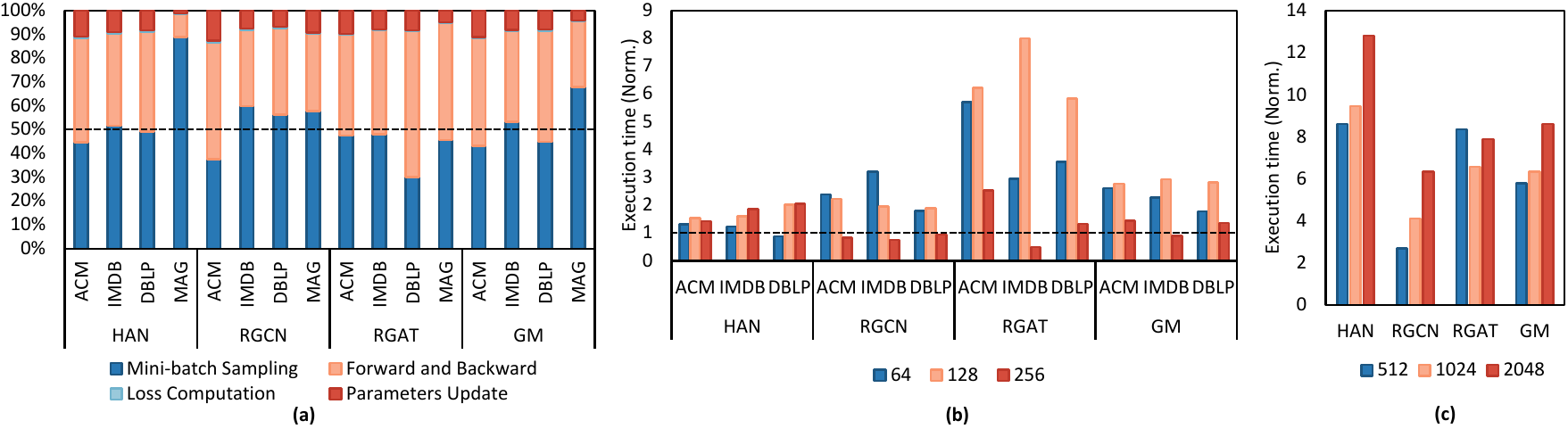}
	\caption{Execution analysis: (a) Time breakdown; (b) Speedup of GPU sampling; (c) Speedup on MAG dataset.}
	\label{fig:single_gpu_mini_batch}
\end{figure*}

\textit{\textbf{Execution Time Breakdown.}} Breaking down the execution time facilitates the identification of the primary execution processes during mini-batch training. Fig.~\ref{fig:single_gpu_mini_batch}(a) shows the breakdown of execution time across various execution stages. \marknumber{15} \textit{Mini-batch sampling occupies the majority of the execution time in each epoch of mini-batch training, even exceeding the combined time of all other execution stages.} This is due to mini-batch sampling in HGNN models requiring sampling operations across multiple semantic graphs not just one graph as in conventional GNNs, entailing traversal of intricate and irregular graph structures. Furthermore, this procedure is carried out on CPU, which is markedly time-intensive in contrast to the highly parallelized computational workload executed on GPU. As presented in Fig.~\ref{fig:single_gpu_mini_batch}(a), mini-batch sampling accounts for an average of 51.01\% of the total time across different models and datasets. Especially for the HAN model on the MAG dataset, the time elapsed on sampling in each epoch accounts for as much as 88.92\%.

\textit{\textbf{Performance Comparison of GPU and CPU Sampling.}} In general, the mini-batch sampling process is categorized as a preprocessing step, managed by the CPU as introduced in Section~\ref{sec:background}. However, compared to CPU sampling, \marknumber{16} \textit{employing GPUs for sampling yields a substantial enhancement in sampling performance, particularly discernible in the context of large-scale datasets}. This is attributed to GPUs offering ample bandwidth resources for graph traversal, coupled with their inherently highly parallel architecture that facilitates simultaneous sampling of multiple target vertices. Fig.~\ref{fig:single_gpu_mini_batch}(b) demonstrates that using GPU for mini-batch sampling directly can achieve an average acceleration of 2.43$\times$, 2.06$\times$, and 1.21$\times$ compared to CPU, under different batch sizes of 64, 128, and 256, respectively. For larger-scale datasets like MAG, GPU sampling achieves an average speedup of 5.78$\times$, 6.35$\times$, and 8.62$\times$ with batch sizes of 512, 1024, and 2048, respectively, as shown in Fig.~\ref{fig:single_gpu_mini_batch}(c).

\textit{\textbf{Characteristics Analysis of the Sampling Process.}} We conduct an in-depth analysis of the mini-batch sampling phase utilizing GPU-based sampling techniques, and present the performance metrics of the primary kernels in Table~\ref{tab:kernel_profiling_sampling}. The performance metrics detailed here pertain exclusively to the CUDA kernels specific to the mini-batch sampling stage, distinguishing them from those involved in the main execution stages.

\begin{table*}[!th]
    \caption{Profiling results of mini-batch sampling of HAN model on MAG dataset.} 
    \label{tab:kernel_profiling_sampling}
    \centering
    \setlength\tabcolsep{5pt}%
	\renewcommand\arraystretch{0.9}
    \resizebox{0.96\textwidth}{!}{
\centering

\begin{tabular}{cccccccc}
\toprule
  \begin{tabular}[c]{@{}c@{}} Kernel \\ Name  \end{tabular}  & 
  \begin{tabular}[c]{@{}c@{}} Time \\ (\%) \end{tabular} &
  \begin{tabular}[c]{@{}c@{}} Achieved Peak \\ Performance (\%) \end{tabular} & 
  \begin{tabular}[c]{@{}c@{}} DRAM BW \\ Utilization (\%) \end{tabular}  &
  \begin{tabular}[c]{@{}c@{}} Shared Memory BW \\ Utilization (\%) \end{tabular}  &
\begin{tabular}[c]{@{}c@{}} L2 Cache \\ Hit  Rate (\%) \end{tabular} &
\begin{tabular}[c]{@{}c@{}} Integer \\ Instructions (Amount) \end{tabular} &
\begin{tabular}[c]{@{}c@{}} Float \\ Instructions (Amount) \end{tabular}\\ \midrule \midrule
    pick\_data & 11.11\% & 0.21\% & 0.22\% & 0.00\% & 91.28\%  & 793156 & 0  \\
    RandomWalkKernel &4.54\% & 17.26\% & 34.47\% & 0.06\% & 86.92\%  & 901362365 & 18032000  \\
    count\_frequency & 36.72\% & 9.16\% & 0.82\% & 0.22\% & 94.64\%  & 457740460 & 0  \\
    compact\_frequency & 1.78\% & 15.57\% & 16.89\% & 8.69\% & 43.61\%  & 165723693 & 0  \\
    count\_hashmap & 0.26\% & 6.92\% & 32.11\% & 1.22\% & 56.42\%  & 19747489 & 593850  \\
    compact\_hashmap & 0.34\% & 5.98\% & 21.94\% & 2.31\% & 67.7\%  & 21827916 & 476162  \\
    RangeKernel & 0.35\% & 11.45\% & 0.00\% & 1.55\% & 100.00\%  & 31911248 & 0  \\
    DeviceSegmentedRadixSortKernel & 26.92\% & 29.56\% & 1.43\% & 42.56\% & 66.39\%  & 5815247002 & 0  \\

    \bottomrule
\end{tabular}
}
\end{table*}

The kernels employed for sampling demonstrate modest computational demands, primarily executing integer instructions for tasks such as vertex ID comparison and indexing. Although the DRAM bandwidth utilization may appear low, this is attributed to the substantial bandwidth provided by the GPU's HBM memory, which is 2039 GB/s on our platform. For instance, the \textit{RandomWalkKernel} achieves an absolute bandwidth value of 702.84 GB/s, significantly exceeding the maximum bandwidth attainable with the CPU's DDR4 memory. The relatively low bandwidth utilization can be attributed to the substantial on-chip cache available on the GPU, which is capable of storing a significant portion of the graph data during the sampling process. This is particularly reflected in the high L2 Cache hit rate, as shown in Table~\ref{tab:kernel_profiling_sampling}.

\subsubsection{Exploring Metapath Changes}
\label{sec:sampling_explore}
In this section, we explore the impact of changes in metapath properties on the efficiency of mini-batch sampling, considering both the length and number of metapaths. We conduct experiments using two datasets, DBLP and MAG, to reflect the varying impacts of metapath property changes across different dataset scales.

\begin{figure*}[!ht] 
	\centering
	\includegraphics[width=1.0\textwidth]{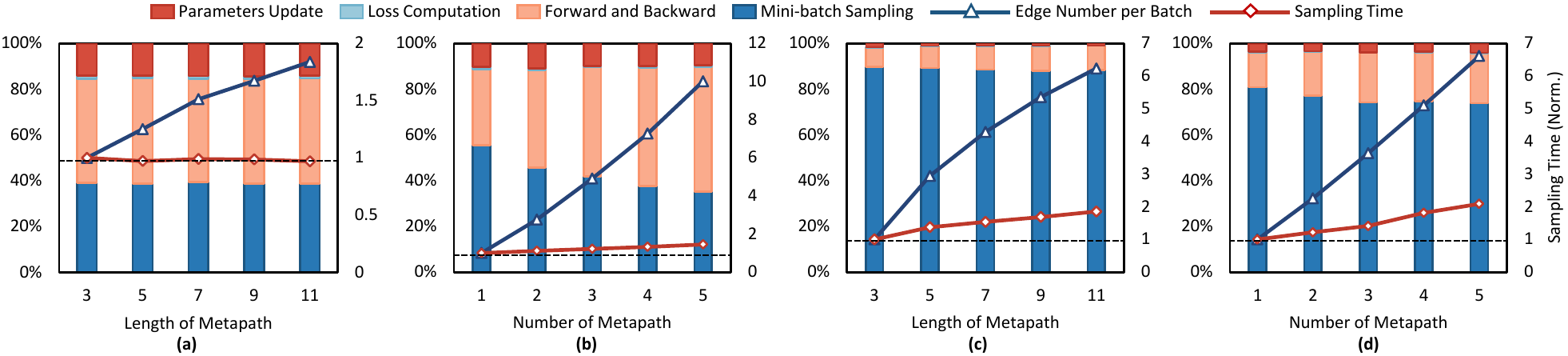}
	\caption{Treands as metapath changes: (a) DBLP dataset; (b) MAG dataset; (c) DBLP dataset; (d) MAG dataset.}
	\label{fig:single_gpu_mini_batch_metapath_changes}
\end{figure*}

\textit{\textbf{Increase in Length of Metapath.}} Fig.~\ref{fig:single_gpu_mini_batch_metapath_changes}(a) and Fig.~\ref{fig:single_gpu_mini_batch_metapath_changes}(b) illustrate the impact on the execution time of mini-batch sampling as metapath length varies. \marknumber{17} \textit{For smaller datasets, the time required for mini-batch sampling exhibits minimal sensitivity to variations in metapath length. Conversely, large-scale datasets experience prolonged execution times as metapath length increases.} Increasing the length of metapath augments the number of edges within the semantic graph, consequently elevating the number of edges sampled per batch. However, thread-level parallelism is utilized during the sampling process. The additional sampling workload, facilitated by adequate hardware resources, does not significantly affect overall performance. Notably, sampling time increases only when the workload surpasses the hardware's maximum parallel processing capacity. As shown in Fig.~\ref{fig:single_gpu_mini_batch_metapath_changes}(a), compared to a metapath length of 3, the sampling time ratio remains around 1 as the metapath length increases on DBLP dataset. However, as depicted in Fig.~\ref{fig:single_gpu_mini_batch_metapath_changes}(b), for larger datasets MAG, as the metapath length increases, the number of sampled edges in each batch grows more significantly, and the mini-batch sampling time also increases accordingly. 



\textit{\textbf{Increase in Number of Metapaths.}} Increasing the number of metapaths directly leads to an increase in the number of semantic graphs. Fig.~\ref{fig:single_gpu_mini_batch_metapath_changes}(c) and Fig.~\ref{fig:single_gpu_mini_batch_metapath_changes}(d) illustrate the variation in mini-batch sampling time with an increasing number of metapaths. \marknumber{18} \textit{For both small-scale and large-scale datasets, the sampling time of mini-batches is sensitive to changes in the number of metapaths.} The increase in the number of semantic graphs due to the growth in metapath number results in an increased sampling load. However, unlike the increase in sampling load caused by denser semantic graphs due to longer metapaths discussed earlier, the sampling process between semantic graphs can only proceed sequentially and is difficult to parallelize within existing programming frameworks. Therefore, even for small datasets, there is a noticeable increasing trend in sampling time as shown in Fig.~\ref{fig:single_gpu_mini_batch_metapath_changes}(c) and Fig.~\ref{fig:single_gpu_mini_batch_metapath_changes}(d).

\section{Multi-GPU Training}
\label{sec:multi}

Distributed training, leveraging multiple GPUs for model training, is driven by the increasing complexity of real-world datasets and models, which can be implemented using either full-batch or mini-batch methods. In comparison to full-batch distributed training, mini-batch distributed training can markedly reduce the convergence time of the training process while preserving model accuracy. Furthermore, it demands fewer hardware resources, rendering it superior in terms of both performance and energy efficiency. Consequently, we focus exclusively on the analysis of the mini-batch distributed training method in this section.

\subsection{Overall Profiling Results}

\subsubsection{Performance Comparison with Different Number of GPUs} 
\begin{figure*}[!htbp] 
	\centering
	\includegraphics[width=0.96\textwidth]{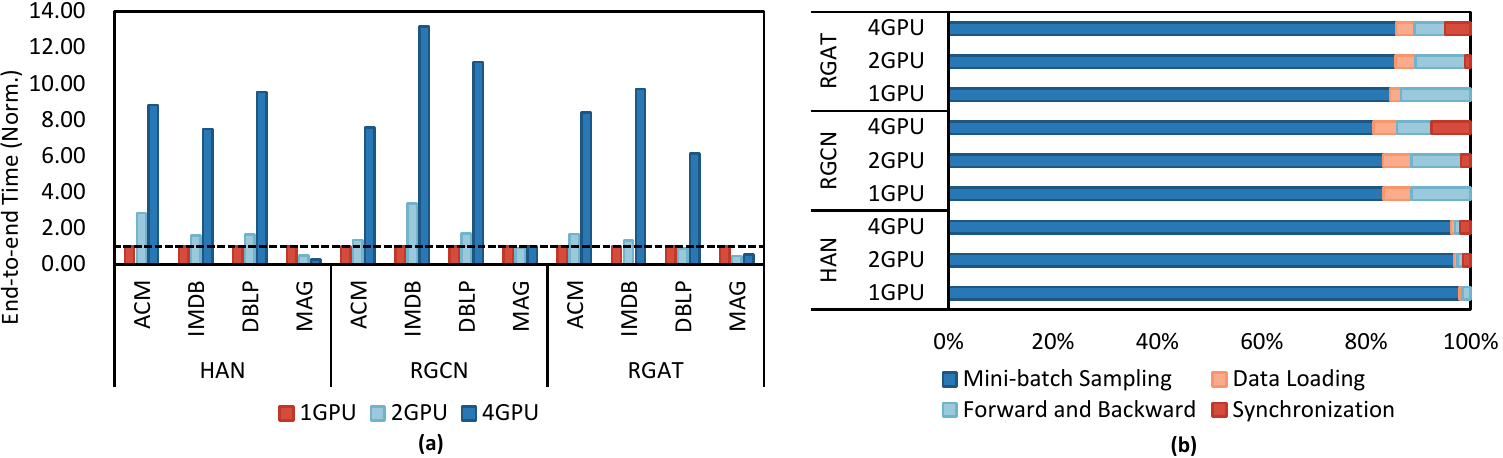}
	\caption{Overall results: (a) Normalized performance compared to training on one GPU; (b) Execution time breakdown on MAG dataset.}
	\label{fig:multi_gpu_perf_breakdown}
	\vspace{-12pt}
\end{figure*}

Fig.~\ref{fig:multi_gpu_perf_breakdown}(a) illustrates the normalized execution time of distributed training using multiple GPUs compared to training on a single GPU. As depicted in the figure, \marknumber{19} \textit{the performance improvement ratio in multi-GPU distributed training scenarios diverges from the ideal ratio.} For smaller-scale datasets such as DBLP, increasing the number of devices may even result in a significant decline in overall training performance. This is mainly because an increase in the number of devices leads to contention for shared resources such as cache and bandwidth, thereby reducing performance. Relevant analysis will be provided in Section~\ref{sec:multi_gpu_analysis}. As shown in Fig.~\ref{fig:multi_gpu_perf_breakdown}(a), the overall execution time of two-GPUs and four-GPUs training are an average of 1.37$\times$ and 8.69$\times$ compared to training on a single GPU for DBLP dataset. While linear improvements are evident in larger-scale datasets like MAG, scenarios involving two GPUs and four GPUs yield average performance enhancements of merely 1.66$\times$ and 2.13$\times$ respectively, falling short of the anticipated 2$\times$ and 4$\times$ improvements. 

\subsubsection{Execution Time Breakdown} Fig.~\ref{fig:multi_gpu_perf_breakdown}(b) presents the execution time breakdown for distributed training with different numbers of GPUs. For clarity, we only present results on the MAG dataset. According to the figure, \marknumber{20} \textit{mini-batch sampling is the predominant execution stage during the distributed training process of HGNN, occupying the vast majority of the execution time.} This is consistent with the mini-batch training scenario on a single GPU, as discussed in Section~\ref{sec:single_gpu_minibatch_execution_analysis}. Due to the complexity of HGNN sampling and the inefficiency of sampling on the CPU, mini-batch sampling has become the primary performance bottleneck.

\subsection{In-depth Analysis}
\label{sec:multi_gpu_analysis}


\subsubsection{CPU Resources Contention During Sampling}
\label{sec:sampling_competition}

To better highlight the decrease in sampling efficiency caused by CPU resource contention, we adopt an adaptive batch size approach here. Specifically, for each epoch, we shuffle the order of target vertices and evenly divide them into $num\_dst\_vertices / num\_gpus$ groups according to the number of GPUs. Each group is then sampled in parallel, and the complete batch after sampling is sent to a separate GPU for execution. We only present the relevant profiling results on the HAN model, with the results on the other two models exhibiting similar characteristics.

\begin{figure*}[!ht] 
	\centering
	\includegraphics[width=1.0\textwidth]{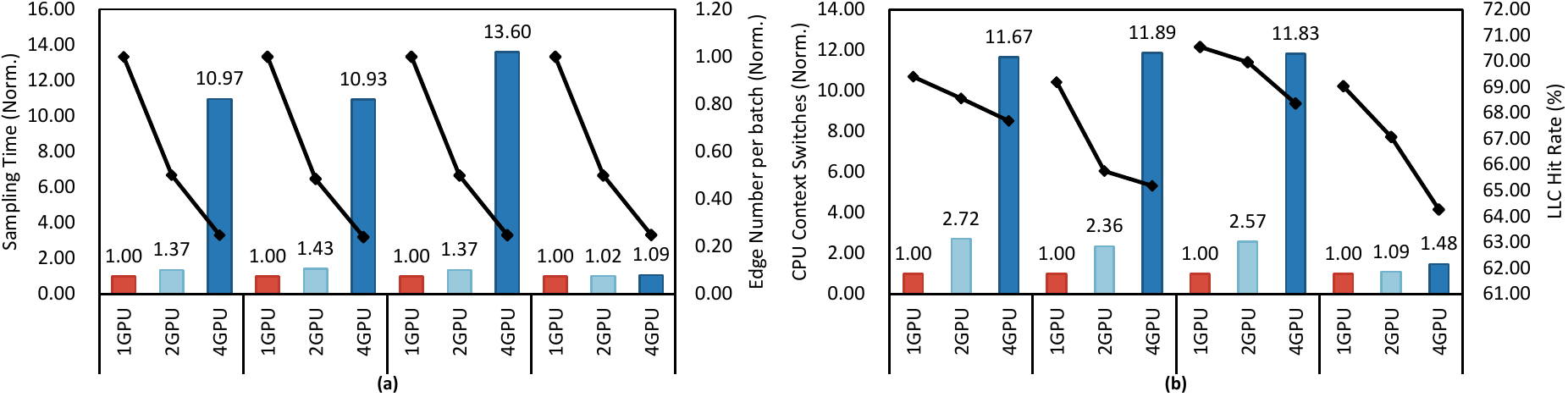}
	\caption{CPU resources contention during mini-batch sampling: (a) Average number of edges in one batch and related sampling time; (b) Number of CPU context switches and LLC hit rate.}
	\label{fig:multi_gpu_sampling_competition}
\end{figure*}

As shown in Fig.~\ref{fig:multi_gpu_sampling_competition}(a), as the number of devices increases, the average number of edges sampled per batch for each GPU decreases. In theory, the average sampling time should also decrease, but in reality, the sampling time shows an increasing trend. Compared to a single GPU, in scenarios with two GPUs and four GPUs, the average time for mini-batch sampling is 1.29 $\times$ and 6.49 $\times$ that of a single GPU, respectively. \marknumber{21} \textit{The primary factor limiting the performance of parallel sampling is the contention for shared CPU resources.} As the number of GPUs increases, the concurrent execution of multiple sampling tasks in thread-level leads to finer partitioning of CPU cores in terms of time slice allocation, resulting in a higher frequency of context switching. As shown in Fig.~\ref{fig:multi_gpu_sampling_competition}(b), when the number of GPUs increases from 1 to 4, the number of CPU context switches increases by an average of 11.79$\times$. This process not only involves operations such as saving and restoring register states but also requires the operating system to schedule tasks based on their priorities. The cumulative delays introduced by these factors contribute to a significant decline in sampling performance. Moreover, the decrease in LLC hit rate suggests that contention among multiple sampling tasks for shared cache resources causes frequent data replacement within the cache, thereby increasing memory access latency and further exacerbating sampling delays.

\subsubsection{Bandwidth Contention During Data Loading and Synchronization} 
\begin{figure*}[!ht] 
	\centering
	\includegraphics[width=0.96\textwidth]{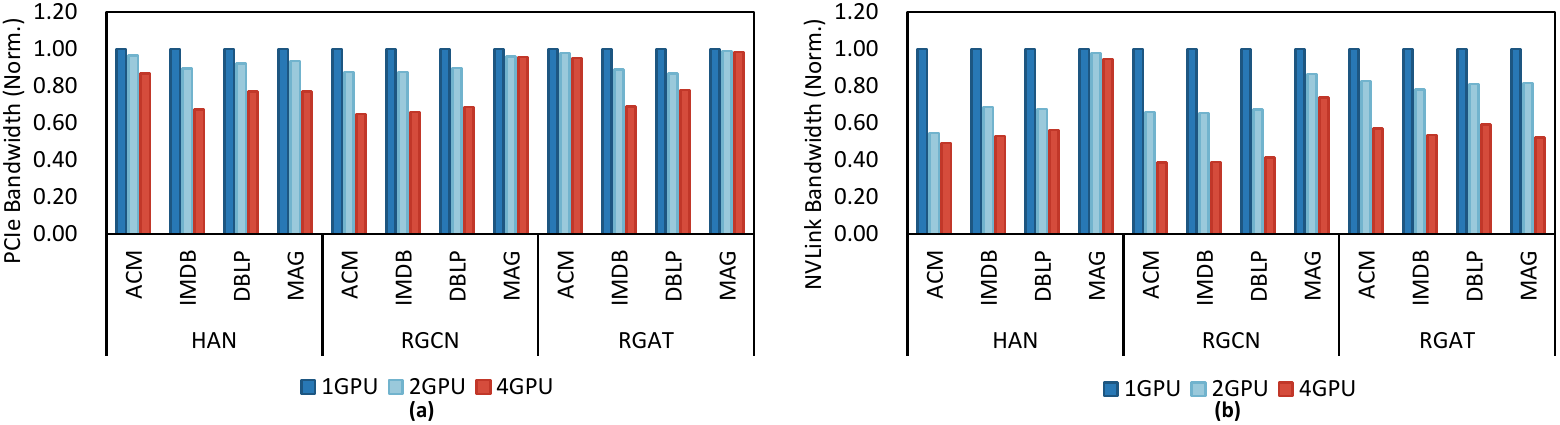}
	\caption{Bandwidth contention between devices: (a) Host-Device PCIe bandwidth; (b) Device-Device NVLink bandwidth.}
	\label{fig:multi_gpu_bandwidth_competition}
\end{figure*}

Data loading and gradient synchronization are two critical steps in distributed training. \marknumber{22} \textit{The factors limiting the performance of data loading and synchronization in multi-device scenarios are primarily the contention for shared bandwidth.} Fig.~\ref{fig:multi_gpu_bandwidth_competition}(a) illustrates the trend of PCIe bus communication bandwidth between the host CPU and GPU devices during distributed training as the number of devices increases. Based on the geometric mean across various model and dataset scenarios, an increase in the number of devices from one to two results in an 8.37\% reduction in the average PCIe bus bandwidth connecting the host and devices. This reduction further extends to 22.62\% when the number of devices increases to four. The decline in PCIe bandwidth suggests that data transfers between the CPU and GPU, particularly the data loading process, encounter increased latency, consequently diminishing the overall performance of distributed training.

Fig.~\ref{fig:multi_gpu_bandwidth_competition}(b) depicts the variation in data transfer bandwidth via NVLink between GPUs during distributed training as the number of devices increases. Based on the geometric mean across various model and dataset configurations, when the number of devices increases from one to two, the average NVLink bandwidth between devices decreases by 26.07\%. When the number of devices increases to four, the average NVLink bandwidth further decreases by 46.03\%. This reduction can be attributed to the use of the NVIDIA Collective Communications Library for gradient synchronization, which, in this study, employs the \textit{All-Reduce} algorithm to synchronize gradients across devices. As this algorithm is executed concurrently on multiple devices, the complexity of the communication network increases with the number of devices, leading to intensified contention for available bandwidth among the devices.





\section{Comparison with GNN Training}
\label{sec:comparison}


In this section, we mainly discuss the comparison between HGNN training and GNN training, focusing on the differences in their execution stages and the transfer of performance bottlenecks during the training process. 

\subsection{Difference in Execution Process}

\subsubsection{Unique Process of Metapath Instance Generation}
\label{sec:SGB}
Given that HGNNs primarily operate on semantic graphs during execution, it is imperative to construct these semantic graphs from the original HetG, which is unique to HGNNs. Our experimental results for the HAN model on different datasets indicate that the execution time of the SGB stage is, on average, 11.46$\times$ the execution time of one single training epoch. Furthermore, as the length and number of metapaths increase, the execution time of the SGB stage exhibits a marked increase, irrespective of the dataset's scale. For large-scale HetG datasets, SGB will be a considerably time-consuming stage during the training process.



\subsubsection{Separate Feature Projection}
\label{sec:gnn_comparison_seprate_FP}

The raw feature vectors of vertices in HomoGs reside in the same vector space and share identical dimensionality, allowing for joint projection. In contrast, HGNNs utilize separate feature projection matrices for each vertex type or each semantic graph, necessitating the reduction of gradients of the projection weight matrices for the same type of vertices appearing in different semantic graphs. As shown in Table~\ref{tab:kernel_profiling_details}, during backward propagation in the FP stage, the percentage of time occupied by the \textit{sgemm} kernel decreased from 98.94\% to 45.3\%, while the \textit{EleWise} kernel increased from less than 1\% to 36.76\%. This shift causes the FP stage in backward propagation to transition from being initially compute-bound to becoming memory-bound, further exacerbating the hybrid execution patterns during the training process.

\subsubsection{Intricate Two-level Aggregation}

GNNs perform a single aggregation step for neighboring vertices within a singular type of relation. In contrast, HGNNs aggregate features from neighbors in each semantic graph generated according to corresponding semantics (relations or metapaths), and then fuse intermediate results of each semantic graph for each vertex. As depicted in Fig.~\ref{fig:single_gpu_time_breakdown_by_phase}, the incorporation of a secondary aggregation level, denoted as the SF stage, incurs an approximate 15.38\% increase in time overhead per epoch.

\subsection{Changes in Execution Bottlenecks}

\subsubsection{More Intricate Hybrid Execution Pattern}

Prior work~\cite{understand_GCN} highlights the presence of a hybrid execution pattern in the execution of the typical GCN model. Specifically, the FP stage (\textit{Combination} stage in work~\cite{understand_GCN}) exhibits a more regular pattern and demonstrates compute-bound execution modes, whereas the NA (\textit{Aggregation} stage) stage involves numerous random accesses, resulting in memory-bound behavior. According to the analysis in Section~\ref{sec:single_gpu_fullbatch_execution_bound_analysis}, compared to GNNs, HGNNs exhibit more pronounced and complex hybrid execution pattern, which stems from their more intricate model structure.

\subsubsection{Bottleneck Migration in Distributed Training}

%
Prior work~\cite{understand_distributed_gnn} demonstrates that the data loading stage dominates the most execution time of GNN distributed training. Conversely, in HGNNs, the mini-batch sampling process incurs markedly higher overhead than data loading and is considered the predominant execution process. This shift arises primarily from the necessity in HGNNs to sample multiple semantic graphs to form a batch, compared to GNNs which sample from a single graph as surveyed in work~\cite{gnn_survey_sampling, survey_graph_processing}. Moreover, models employing metapath-based graph construction experience significant overhead in neighbor sampling due to the intricate traversal of multi-hop neighbors following each meatapath.


\section{Optimization Guidelines}
\label{sec:guidelines}

In this section, leveraging the findings previously outlined, we provide guidance for optimizing HGNN training from both software and hardware perspectives.

\subsection{Software Optimizations}

\subsubsection{Reasonable Overlapping of Phases} On one hand, a bound-aware kernel fusion method can be proposed to facilitate the overlapping execution of stages with differing bounds. Observations \marknumber{5} and \marknumber{6} underscore the presence of intricate hybrid execution patterns during HGNN training. These stages, distinguished by diverse execution bounds, frequently alternate, facilitating overlap execution to harness multiple hardware resources concurrently. This approach enhances overall execution performance by optimizing hardware resource utilization. For example, \textit{Graphite}~\cite{Graphite} adopts a similar approach to accelerate GNNs on the CPU.

Moreover, training phases executed on different devices can be overlapped to reduce overall time overhead. Observation \marknumber{15} and \marknumber{20} indicate that using a mini-batch-based training method makes the mini-batch sampling process the primary execution stage. Fortunately, there is inherent parallelism between the sampling process executed on the CPU and the workload computation process executed on the GPU. The training paradigm can be adjusted to start the sampling process for the next epoch while performing the computation for the current epoch, like \textit{PaGraph}~\cite{PaGraph} which overlaps mini-batch sampling with data loading to eliminate sampling overhead during GNN training.

\subsubsection{Recomputing to Reduce Memory Cost}

Observation \marknumber{7} and \marknumber{9} indicate that compared to the inference process alone, the training process requires more memory storage, primarily due to the need for direct memory access to reuse a large number of intermediate results preserved during the forward propagation. However, observation \marknumber{4} suggests that operations involving addition in the forward propagation process incur no computational cost in the backward propagation process. This characteristic of backward propagation, reducing computation while increasing memory usage compared to forward propagation, enables the possibility of recomputing certain intermediate results to conserve memory. Prior work~\cite{mlsys2022recomputation} proposes an evaluation method to strike a reasonable trade-off between recomputation cost and memory overhead, aiming to judiciously recompute certain intermediate variables during backward propagation to reduce storage expenses in GNN training. Additionally, \textit{TT-GNN}~\cite{TT-GNN} also employs a method of recomputing prefix arrays to reduce memory cost during GNN training.

\subsubsection{Scheduling Based on Semantic Graphs for Data Reuse}

Processing multiple semantic graphs is an important feature of HGNNs compared to GNNs. Determining the optimal execution order of semantic graphs based on the number of shared vertices between different semantic graphs can maximize data reuse between the graphs, thereby reducing off-chip memory accesses. Pioneered by \textit{HiHGNN}~\cite{HiHGNN}, the concept of semantic graph similarity was introduced to maximize reusable data in HGNN inference processes. This can also be extended and applied to HGNN training, exploring optimal execution orders across multiple layers and epochs to maximize data reuse and reduce off-chip memory accesses. Besides, \textit{GDR-HGNN}\cite{GDR-HGNN} utilizes the bipartite nature of semantic graphs to decompose them in order to enhance data locality.

\subsection{Hardware Optimizations}


\subsubsection{Independent Parallel Neighbor Traversal Unit} Both the extremely time-consuming mini-batch sampling and SGB stages involve traversing neighbors of target vertices. As indicated by the analysis in Section~\ref{sec:sampling_explore}, this traversal process can be effectively parallelized across various target vertices. However, Section~\ref{sec:sampling_competition} and observation \marknumber{21} underscore that concurrent sampling processes frequently contend for shared resources, leading to notable performance degradation in neighbor sampling tasks. Consequently, it is advisable for researchers to design specialized neighbor traversal units. Each unit should independently cache relevant neighbor information specific to its assigned target vertex, thereby leveraging sampling parallelism while mitigating cache contention. Moreover, optimization structures tailored for graph processing with irregular memory access patterns such as \textit{Graphicinado}~\cite{Graphicionado} and \textit{GraphDynS}~\cite{GraphDynS} could also be contemplated for enhancing neighbor traversal efficiency.

\subsubsection{Unified Reconfigurable Execution Unit}

Observations \marknumber{5} and \marknumber{6} underscore the presence of intricate hybrid execution patterns during HGNN training, resulting in varied utilization of hardware resources across different stages and an inability to fully harness the maximum efficiency of the hardware platform. \textit{ADE-HGNN}~\cite{ADE-HGNN} advocates for employing reconfigurable architectures to streamline HGNN inference execution. Given the complex nature of the training process, unified architectures hold the potential to substantially optimize hardware resource utilization.

\subsubsection{Reuse-distance-aware Cache}
Finding \marknumber{11} reveals that memory dependency is the primary cause of execution stalls in most CUDA kernels during HGNN training, with memory access latency serving as a significant performance bottleneck. Additionally, observation \marknumber{7} shows that backward propagation incurs more DRAM accesses than forward propagation, while finding \marknumber{10} reveals a lower cache hit rate during backward passes. This is due to the irregularity of graph data, the broader range of accessed data, and the deeper computational graphs in training, resulting in longer reuse distances and frequent eviction to off-chip memory before reuse. From a holistic perspective, researchers can model the computational graph of HGNN training, and, by considering data size and reuse distance as a function of graph depth, compute a comprehensive replacement weight index. This index can guide the design of specialized cache structures that prioritize retaining data with shorter reuse distances, thereby improving cache hit rates and reducing off-chip memory access.

\subsubsection{Multi-lane Architecture Supporting Semantic-graph-level Parallelism}

Besides enabling data reuse, multiple semantic graphs offer a form of parallelism unique to traditional GNNs, termed semantic graph parallelism. During the training process of HGNN models, the operations within each semantic graph prior to the SF stage are independent of other graphs, inherently possessing parallelism. Researchers can improve training efficiency by devising multi-lane hardware architectures that facilitate the concurrent execution of various semantic graphs like \textit{HiHGNN}~\cite{HiHGNN} proposed for HGNN inference acceleration.

\section{Conclusion}
\label{sec:conclusion}

The complex and costly training process is crucial for effectively utilizing HGNNs. In this work, we comprehensively analyze different training methods and scenarios using NVIDIA GPU A100 platform, revealing the execution semantics and patterns of the HGNN training process, and uncovering the performance bottlenecks. Additionally, we compare some similarities and differences between HGNN and GNN training. Finally, we provide optimization guidelines from both software and hardware perspectives.

\bibliographystyle{ACM-Reference-Format}
\bibliography{references.bib}

\end{document}